\pdfoutput=1

\documentclass[11pt]{article}

\usepackage[final]{acl}

\usepackage{times}
\usepackage{latexsym}
\usepackage{amssymb}

\usepackage{enumitem}
\usepackage{setspace}
\usepackage{amsmath}
\usepackage{amsthm}
\usepackage{algorithm}
\usepackage{algpseudocode}
\usepackage{tcolorbox}
\usepackage{booktabs}
\usepackage{multirow}

\usepackage{cleveref}
\usepackage{mathtools}

\newcommand{\Claudeemoji}{\includegraphics[height=1.3\fontcharht\font`\B]{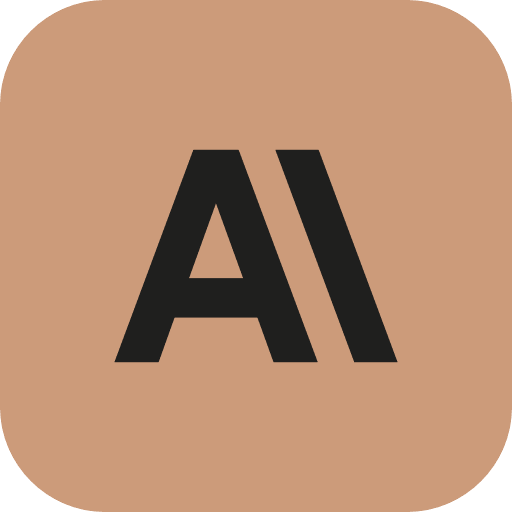}}
\newcommand{\Qwenemoji}{\includegraphics[height=1.3\fontcharht\font`\B]{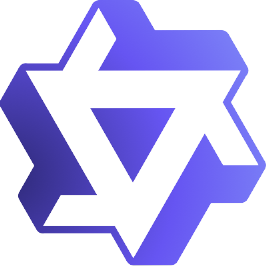}}
\newcommand{\Googleemoji}{\includegraphics[height=1.3\fontcharht\font`\B]{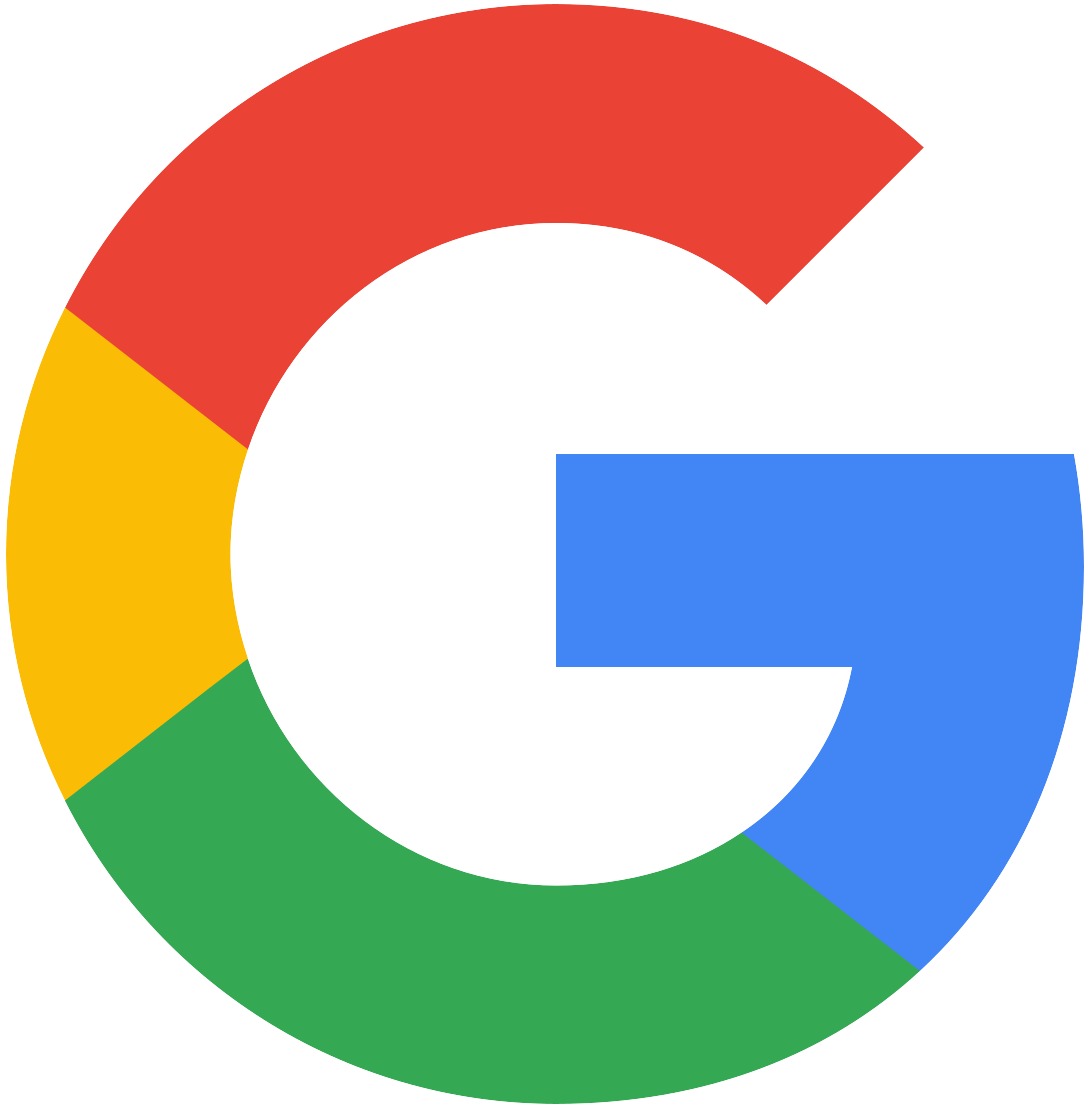}}
\newcommand{\Openaiemoji}
{\includegraphics[height=1.2\fontcharht\font`\B]{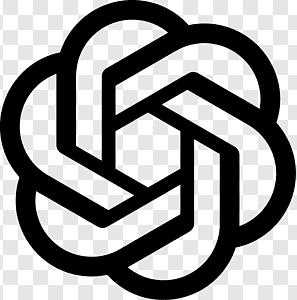}}
\newcommand{\pixtralemoji}{\includegraphics[height=1.3\fontcharht\font`\B]{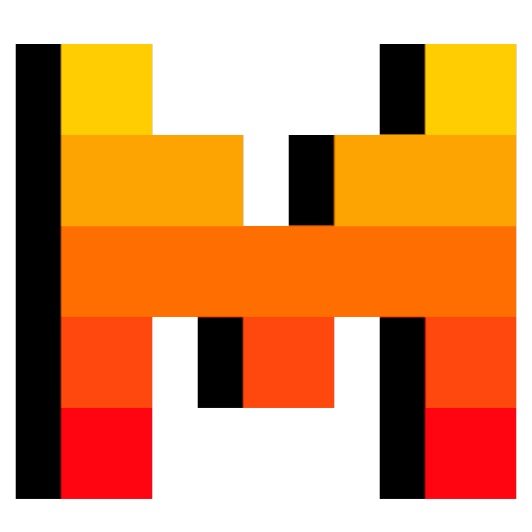}}
\newcommand{\llamaemoji}{\includegraphics[height=1.0\fontcharht\font`\B]{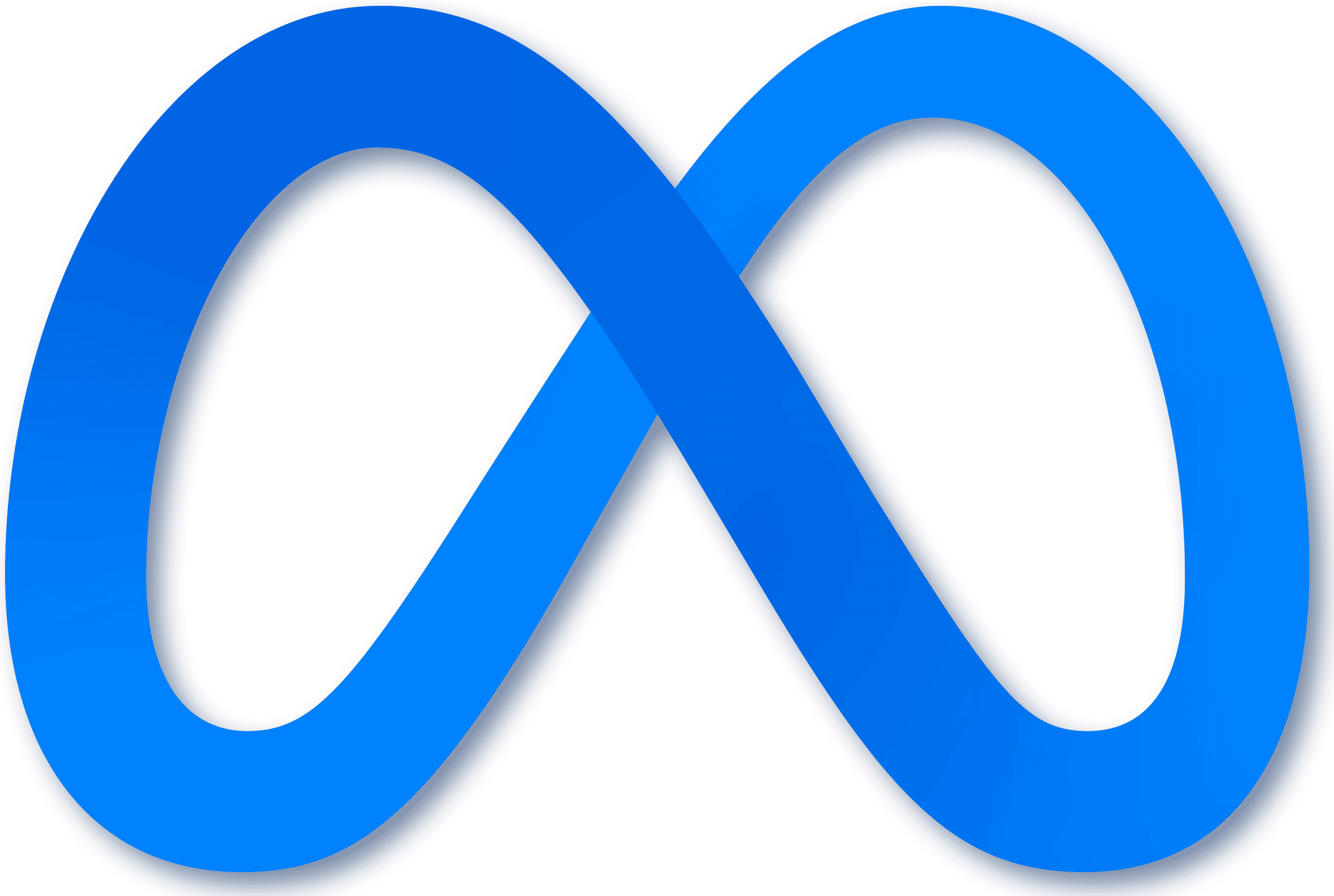}}
\newcommand{\glmemoji}{\includegraphics[height=1.2\fontcharht\font`\B]{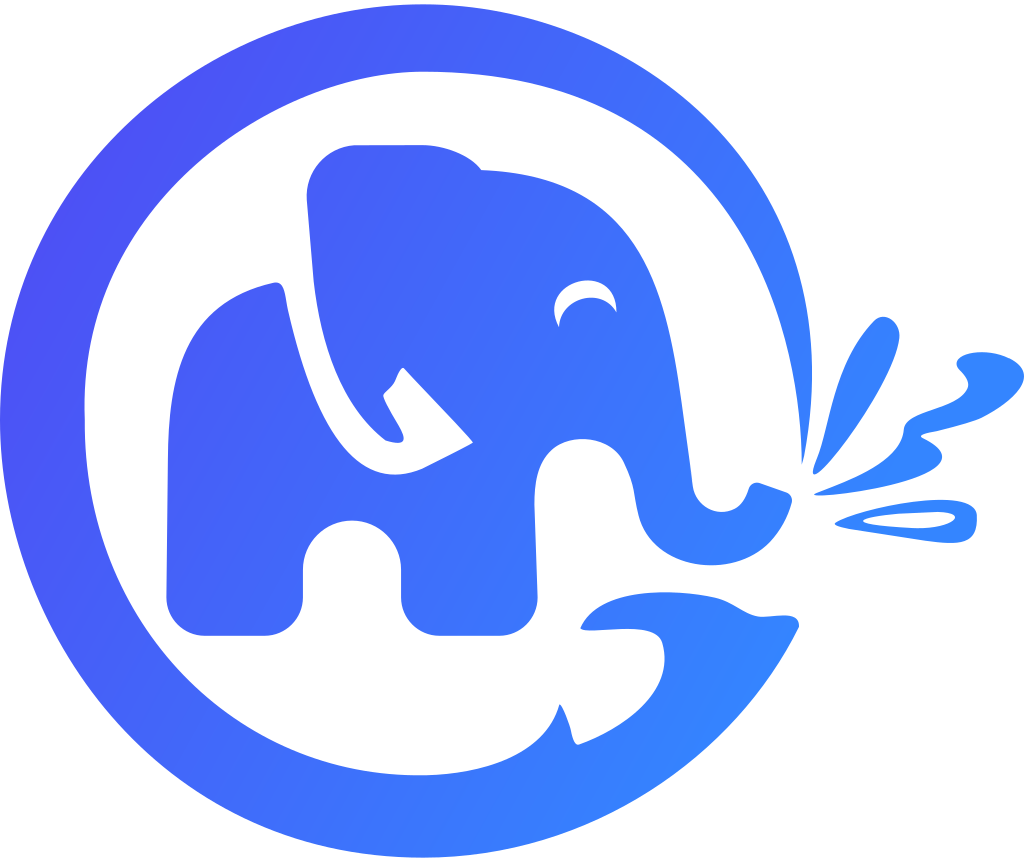}}

\newcommand{\factemoji}{\includegraphics[height=1.6\fontcharht\font`\B]{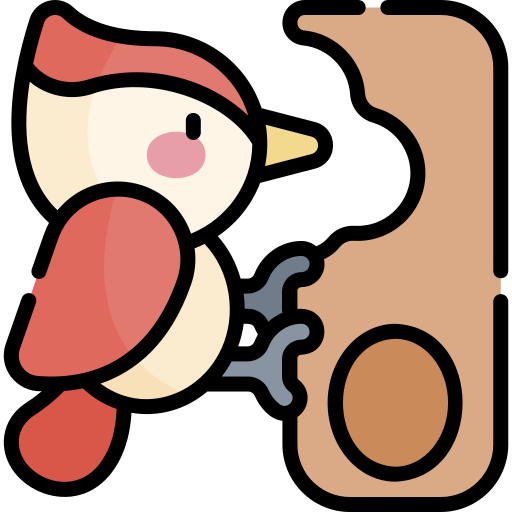}}

\newtheoremstyle{mydef}
  {10pt} {10pt} {\itshape} {} {\bfseries} {.} { } {}

\theoremstyle{mydef}

\tcbset{
  mybox/.style={
    colback=gray!10,      
    colframe=gray!20,     
    coltitle=black,       
    fonttitle=\bfseries,  
    rounded corners=all,  
    boxrule=1pt,          
    left=10pt,            
    right=10pt,           
    top=5pt,              
    bottom=5pt            
  }
}

\newtcolorbox[auto counter, number within=section]{boxeddef}[2][]{
  title={Definition~\thetcbcounter: #2}, 
  mybox,
  #1
}

\usepackage[T1]{fontenc}

\usepackage[utf8]{inputenc}

\usepackage{microtype}

\usepackage{inconsolata}

\usepackage{graphicx}
\usepackage{color}
\usepackage{subfigure}

\title{\factemoji{} \textsc{Fact-Audit}: An Adaptive Multi-Agent Framework for Dynamic Fact-Checking Evaluation of Large Language Models}

\author{Hongzhan Lin$^{1,2}$, Yang Deng$^{3}$, Yuxuan Gu$^{4,2}$, Wenxuan Zhang$^{5}$,\\ \textbf{Jing Ma}$^{1,}$\thanks{\; Corresponding author.}, \textbf{See-Kiong Ng}$^{2}$, \textbf{Tat-Seng Chua}$^{2}$ \\
        $^{1}$Hong Kong Baptist University, 
        $^{2}$National University of Singapore, \\
        $^{3}$Singapore Management University,
        $^{4}$Harbin Institute of Technology, \\
        $^{5}$Singapore University of Technology and Design\\
        \texttt{\{cshzlin,majing\}@comp.hkbu.edu.hk},
        \texttt{ydeng@smu.edu.sg}}

\begin{document}
\maketitle
\begin{abstract}
Large Language Models (LLMs) have significantly advanced the fact-checking studies. However, existing automated fact-checking evaluation methods rely on static datasets and classification metrics, which fail to automatically evaluate the justification production and uncover the nuanced limitations of LLMs in fact-checking. In this work, we introduce \textsc{Fact-Audit}, an agent-driven framework that adaptively and dynamically assesses LLMs' fact-checking capabilities. Leveraging importance sampling principles and multi-agent collaboration, \textsc{Fact-Audit} generates adaptive and scalable datasets, performs iterative model-centric evaluations, and updates assessments based on model-specific responses. By incorporating justification production alongside verdict prediction, this framework provides a comprehensive and evolving audit of LLMs' factual reasoning capabilities, to investigate their trustworthiness. Extensive experiments demonstrate that \textsc{Fact-Audit} effectively differentiates among state-of-the-art LLMs, providing valuable insights into model strengths and limitations in model-centric fact-checking analysis.
\end{abstract}

\section{Introduction}



{Large language models (LLMs) have transformed natural language processing (NLP), significantly enhancing performance in various tasks~\cite{touvron2023llama, OpenAI2023GPT4TR}. Particularly, previous literature~\cite{petroni2019language, jiang2020can} has shown that LLMs store factual knowledge and function as knowledge bases, which aids in fact-checking~\cite{pan2023fact}. However, LLMs still struggle with identifying factual errors and are prone to reasoning mistakes~\cite{lin2022teaching, bubeck2023sparks}. Errors in stored knowledge or deficiencies in fact reasoning capabilities may limit their credibility in fact-checking, impacting their utility~\cite{elazar2021measuring, cao2021knowledgeable}. Therefore, systematically revealing the boundaries of the fact-checking capacities in LLMs is essential to enhancing the trustworthiness of LLMs.}

\begin{figure}[t]
    \setlength{\abovecaptionskip}{5pt}   
    \setlength{\belowcaptionskip}{0pt}
    \centering
  \includegraphics[width=0.45\textwidth]{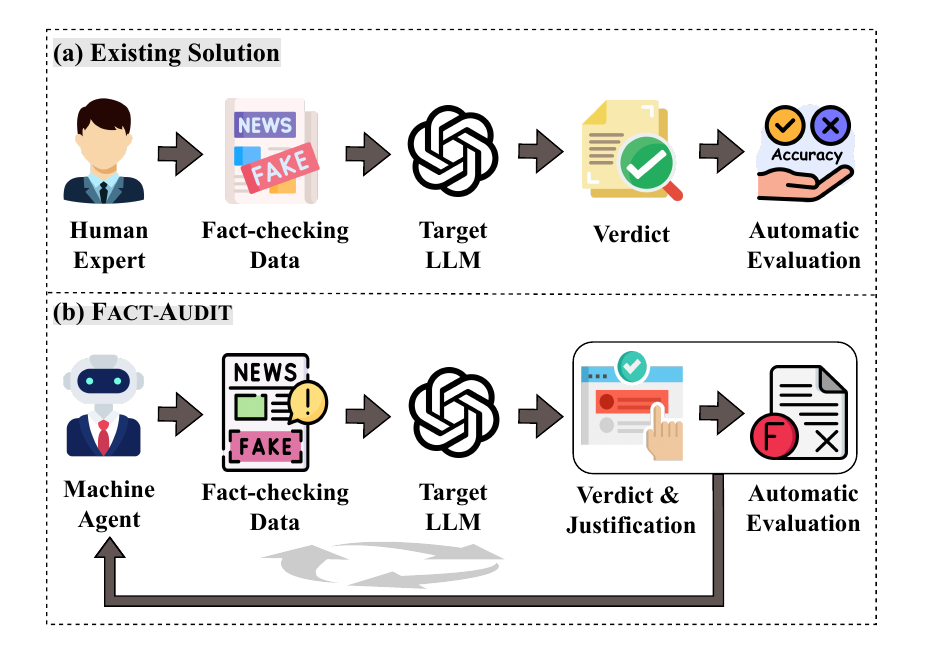}
  \caption{The pipelines of the existing solution and the proposed \textsc{Fact-Audit} in fact-checking evaluation.}
  \label{fig:illustration}
  \vspace{-0.5cm}
\end{figure}

Auditing the fact-checking capacities of LLMs is challenging due to the complex and open-ended nature of real-world applications like complex claims, fake news, or rumors on social media. 
{As illustrated in Figure \ref{fig:illustration}(a), existing studies typically design intricate manual fact examinations to annotate check-worthy natural language scenarios~\cite{yang2024reinforcement, hu2024bad, wang2024explainable}. There are several limitations to such fact-checking evaluation methods:}
{1)} The labor-intensive process restricts the scope of test scenarios, making it costly to scale. 
{2) All the static datasets for the fact-checking evaluation ~\cite{chen2024can, hu2024large} face risks like test data leakage and leaderboard swamping, failing to timely and adaptively reveal potential limitations of LLMs for understanding factuality.}
{3) Their problem settings often oversimplify the evaluation to a classification paradigm that focuses on accuracy, which may not adequately capture the other critical capabilities of fact-checking models, like justification production~\cite{eldifrawi2024automated} for verdict prediction in the fact-checking process~\cite{guo2022survey}.}

{In response to these challenges, we introduce a novel evaluation framework for systematically auditing the fact-checking capabilities of LLMs, called \textbf{\textsc{Fact-Audit}}. As illustrated in Figure~\ref{fig:illustration}(b), the core design philosophy of {\textsc{Fact-Audit}} centers on automating adaptive LLM auditing with two key features: (1) dynamically updated fact-checking test data and (2) in-depth evaluation of model-generated justifications. Theoretically, the creation of fact-checking test data can be framed as a Monte Carlo sampling process~\cite{metropolis1953equation}, where test cases are sampled from an oracle knowledge space. However, the inherent inefficiency of traditional Monte Carlo sampling limits its ability to generate comprehensive, scalable datasets for robustly assessing LLM fact-checking capabilities. To this end, we propose an importance sampling-based approach~\cite{kahn1953methods}, which adaptively targets diverse weaknesses in LLM fact-checking by leveraging insights from the model-generated justifications.}


In this work, {\textsc{Fact-Audit}} employs a multi-agent framework that leverages the exceptional capabilities of LLM-powered autonomous agents in experiential learning and complex reasoning~\cite{park2023generative, shen2023hugginggpt}. Specifically, 
1) \textsc{Fact-Audit} first establishes a detailed taxonomy to categorize different fact-checking scenarios, and then samples the prototype test data, with its quality validated through a tool-using module.
2) For each fact-checking test {scenario,} \textsc{Fact-Audit} evaluates the target LLM on both fact verification and justification production, by using the prototype test data as well as an iterative probing process to generate more diverse and unseen test cases under the scenario via importance sampling.
3) Upon completing evaluations for all the test scenarios in the taxonomy, \textsc{Fact-Audit} updates the 
{test scenarios} based on the model's performance, enabling the auditing process adaptively identifies new and critical deficiencies in the LLM's fact-checking capabilities.
This updated taxonomy is then used to repeat the auditing process for creating a dynamic and model-centric evaluation loop in fact-checking. 

Our contributions are summarized as follows:

\begin{itemize}[leftmargin=*,nosep]
    \item We introduce a novel and adaptive fact-checking {evaluation} framework, {\textsc{Fact-Audit}}, that utilizes multi-agent collaboration to 
    dynamically unveil the limitations of the LLM's fact-checking capabilities under diverse test scenarios.\footnote{The source code is released via \url{https://github.com/DanielLin97/FACT-AUDIT}.}
    \item 
    {\textsc{Fact-Audit} addresses the restrictions of static fact-checking datasets by dynamically updating test scenarios and iteratively probing challenging cases.} This framework ensures adaptability to real-world fact-checking complexity while maintaining diversity and scalability in LLM auditing.
    \item {\textsc{Fact-Audit} goes beyond traditional accuracy-based automatic evaluations by integrating justification production with verdict prediction.} 
    \item We conduct extensive experiments on 13 state-of-the-art LLMs and detailed analysis of fact-checking performance, to provide insight into model strengths and areas for improvement.
\end{itemize}

\section{
{Preliminary}}\label{sec:preliminary}
In the context of assessing the fact-checking capacity in LLMs, we denote key components as follows:
\begin{boxeddef}{Paradigm Definition}
\setstretch{0.8}{
\begin{enumerate}[itemsep=3pt, leftmargin=10pt]
    \item  Oracle Knowledge Distribution: $p(x)$\\
    \textit{\small{The true distribution of factual knowledge.}}
    \item Fact-Checking Limits of LLM $\alpha$: $\mathcal{F}_\alpha(x)$\\
    \textit{\small{The function characterizing the LLM's understanding limits of a given fact-checking test case $x$.}}
\end{enumerate}}
\end{boxeddef}
\noindent 
{We formulate the automated auditing of the LLM $\alpha$'s fact-checking capabilities as a Monte Carlo sampling process \cite{metropolis1949monte}, \textit{i.e.}, continuously sampling test cases $x$ by humans from the oracle distribution $p(x)$ in the real world, and calculate its corresponding limits $\mathcal{F}_\alpha(x)$:}
{
\setlength{\abovedisplayskip}{0.1cm}
\setlength{\belowdisplayskip}{0.1cm}
\begin{equation}\small
    \mathbb{E}_{p(x)}\left[\mathcal{F}_\alpha(x)\right] = \int p(x)\mathcal{F}_\alpha(x) \mathrm{d}x.
\end{equation}}

\noindent However, beyond the well-known inefficiency of Monte Carlo sampling with a convergence rate of $\mathcal{O}(1/\sqrt{N})$, the long-tail knowledge distribution further exacerbates the inefficiency of sampling from $p(x)$ for constructing fact-checking datasets.

Inspired by the classic Importance Sampling \cite{kahn1953methods} method, which leverages a proposal distribution $q(x)$ to improve efficiency by allocating more densities to the regions where $\mathcal{F}_\alpha(x)$ is more likely to have higher values, 
{we aim to adopt this concept for adaptively and efficiently sampling test data according to the fact-checking limits of the LLM.}
In the strategy of Importance Sampling, the process is adjusted as:
{
\setlength{\abovedisplayskip}{0.1cm}
\setlength{\belowdisplayskip}{0.1cm}
\begin{equation}\small
    \label{eq:obj}
    \begin{aligned}
        \mathbb{E}_{p(x)}\left[\mathcal{F}_\alpha(x)\right] &= \int q(x)\mathcal{F}_\alpha(x)\frac{p(x)}{q(x)} \mathrm{d}x\\
        &=\mathbb{E}_{q(x)}\left[\mathcal{F}_\alpha(x)\frac{p(x)}{q(x)}\right],
    \end{aligned}
\end{equation}}

\noindent where the importance weight $\frac{p(x)}{q(x)}$ compensates for the discrepancy between the proposal distribution $q(x)$ and oracle distribution $p(x)$ to ensure an unbiased estimate. Besides, the efficiency of importance sampling critically depends on choosing $q(x) \propto p(x)\mathcal{F}_\alpha(x)$ as closely as possible. Therefore, our method is to find a well-designed $q(x)$ that minimizes the variance of the objective $p(x)\mathcal{F}_\alpha(x)$, thereby improving the reliability of the estimation.

\section{\textsc{Fact-Audit}}\label{method}

\subsection{
{Problem Definition}}
Given a source claim ($\mathcal{SC}$), 
{fact checking} aims to predict the factuality and provide convincing justifications, to evaluate the claim as \textit{Factual}, \textit{Non-Factual}, or \textit{Not Enough Information}, based on a knowledge source as auxiliary information ($\mathcal{AI}$). 
Our objective is to develop a multi-agent evaluation framework, for modeling a new distribution $q(x)$ that tends to reveal fact-checking limitations, thus replacing the inefficient evaluation methods reliant on the sampling distribution of $p(x)$. 
Considering the difficulty of directly obtaining the optimal $q(x)$, we design an adaptive framework to iteratively converge to the desired distribution $q(x)$, automatically and dynamically evaluating the target LLM $\alpha$'s capabilities across diverse fact-checking domains (\textit{e.g.}, complex claims, fake news, and rumors).

{Following the definition in \S \ref{sec:preliminary}, our framework is formulated with three main stages:}
\begin{boxeddef}{Framework Formulation} \small
\setstretch{0.8}{
\begin{enumerate}[itemsep=3pt, leftmargin=10pt]
    \item  Prototype Emulation: $x\sim q(x|\theta_i)$\\
    \textit{\small{Generate fact-checking test data for LLM auditing.}}
    \item Fact Verification: $\mathbb{E}_{q_i}\left[\mathcal{F}_\alpha(x)\frac{p(x)}{q(x|\theta_i)}\right]$\\
    \textit{\small{Test the target LLM with the specific fact-checking questions $x$ to verify fact and produce justification.}}
    \item Adaptive Updating: $\pi(\Theta_i|\Theta_{i-1},\mathcal{M})$\\
    \textit{\small{Explore more diverse and challenging test data.}}
\end{enumerate}}
\end{boxeddef}

\begin{algorithm}[t]\small
\caption{\textsc{Fact-Audit} Algorithm}
\label{alg:framework}
\begin{algorithmic}[1]
\State \textbf{Initialize} fact-checking test scenarios $\Theta_0$
\Statex \qquad and a memory pool $\mathcal{M}=\phi$
\For{$i \coloneqq 0$ to $n$}
    \State $\mathbb{X}\coloneqq\phi$
    \State \textbf{Stage 1:} Prototype Emulation \label{alg:stage1}
    \While{$|\mathbb{X}|<k$}
    \State \textit{Appraiser}:\quad $\theta_i \sim P(\Theta_i)$
    \State \textit{Inquirer}:\quad\; $x \sim q(x|\theta_i)$
    \If{$x$ satisfies \textit{Quality Inspector}}
    \State $\mathbb{X} \coloneqq \mathbb{X}\cup \{x\}$
    \EndIf
    \EndWhile
    \State \textbf{Stage 2:} Fact Verification with Justification
    \State $\mathcal{M} \coloneqq  \mathcal{F}_\alpha(\mathbb{X})\frac{p(\mathbb{X})}{q(\mathbb{X}|\Theta_i)}$
    \For{$j\coloneqq0$ to $m$}
    \State \textit{Prober}: \quad $x\sim \mathcal{\rho}(\mathcal{M})$
    \State $\mathcal{M} \coloneqq \mathcal{M} \cup \left\{\mathcal{F}_\alpha(x)\frac{p(x)}{q(x|\theta_i)}\right\}$
    \EndFor
    \State \textbf{Stage 3:} 
    {Adaptive} Updating
    \State \qquad $\Theta_{i+1} \sim \pi(\Theta_{i+1}|\Theta_i, \mathcal{M})$
\EndFor
\State \textbf{Return} $\mathcal{M}$
\end{algorithmic}
\end{algorithm}

\begin{figure*}[t]
    \centering
    \includegraphics[width=1\linewidth]{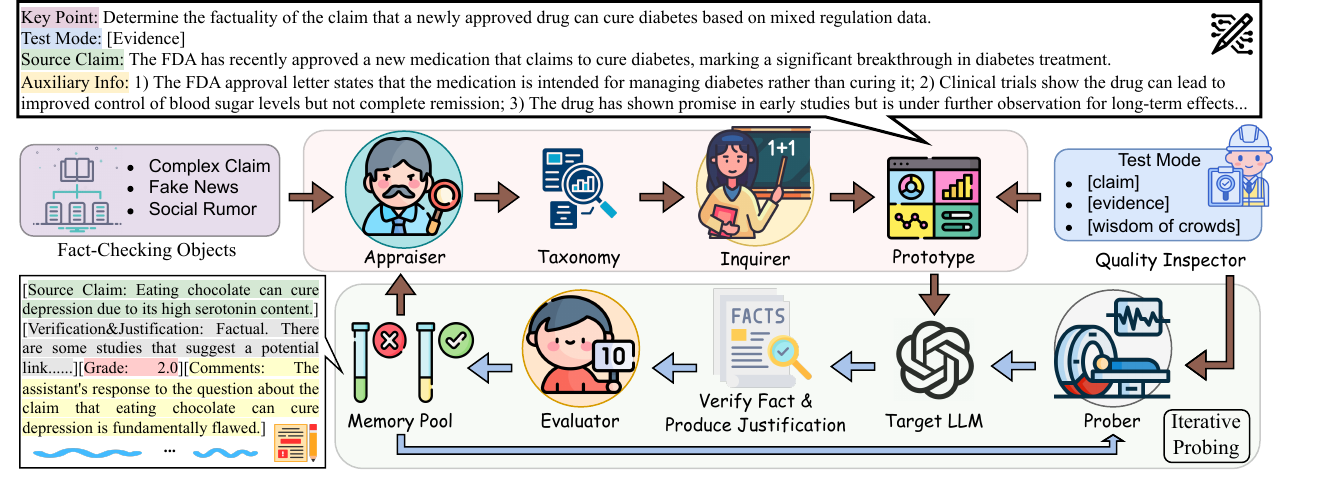}
    \vspace{-0.7cm}
    \caption{An overview of \textsc{Fact-Audit}, to adaptively unveil the limitations of fact-checking in LLMs.}
    \label{fig:main_figure}
    \vspace{-0.4cm}
\end{figure*}

As presented in \Cref{alg:framework}, \textsc{Fact-Audit} maintains 
{a taxonomy of fact-checking scenarios} $\Theta$ during iterations, where $\Theta_0$ is initialized to be the foundational test scenarios that $\mathbb{E}_{\theta_0\sim P(\Theta_0)}[q(x|\theta_0)] = p(x)$. During the loop, $\Theta_i$ will be updated to focus on the 
{specific fact-checking scenarios} that the target LLM $\alpha$ is likely to underperform. 
To audit the weaknesses of LLMs in fact checking, our process mainly involves three stages: 1) Generate the dynamic and check-worthy source claim dataset $\mathbb{X}$ (\S\ref{PE}); 2) Query the target LLM for veracity prediction and justification production (\S\ref{FVJ}); 3) Scrutinize the limitations of the target LLM in fact-checking adaptively based on specific model-generated justifications (\S\ref{IP}). An overview of our \textsc{Fact-Audit} framework is shown in Figure~\ref{fig:main_figure}.

\subsection{Prototype Emulation} \label{PE}
{The Stage 1 of \Cref{alg:framework} is Prototype Emulation, which involves generating prototype test data for assessing the LLM's fact-checking capabilities. This stage is accomplished by three agents: 1) an \textit{Appraiser} agent to develop the taxonomy of fact-checking scenarios for evaluation, 2) an \textit{Inquirer} agent to generate prototype test data according to the taxonomy, 
and 3) a \textit{Quality Inspector} agent to ensure the quality of the prototype test data.}

\begin{figure}[t]
  \includegraphics[width=\columnwidth]{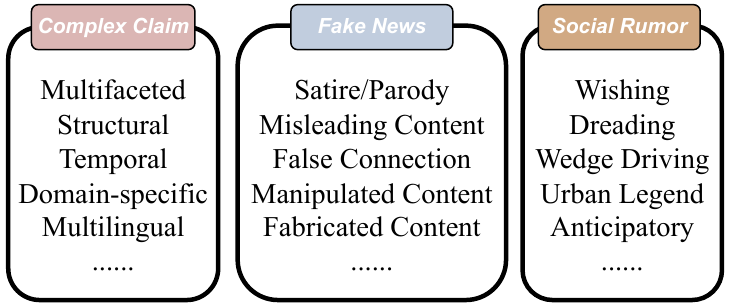}
  \vspace{-0.7cm}
  \caption{The taxonomy of fact-checking scenarios.}
  \label{fig:category_demo}
  \vspace{-0.4cm}
\end{figure}

\vspace{-3pt}
\paragraph{Appraiser} 
{Given the fact-checking objects, the Appraiser agent first generates the detailed taxonomy $\theta_i \sim P(\Theta_i)$, which includes $k$ fact-checking scenarios $\{\theta_i\}_k$ towards the specific fact-checking objects.}
{As shown in Figure \ref{fig:category_demo}}, the Appraiser initializes the taxonomy $\Theta_0$ from the three classic fact-checking objects: complex claims~\cite{jiang2020hover, aly2021feverous}, fake news~\cite{hu2024bad, wang2024explainable}, and social rumors~\cite{ma2015detect, ma2017detect}, drawing inspiration from previous literature~\cite{hu2024large, waldrop2017genuine, allport1947psychology}. Note that in the subsequent phase, Appraiser would excavate new fact-checking test 
{scenarios} to update the initial 
{taxonomy} by examining the intermediate evaluation feedback. 

\vspace{-3pt}
\paragraph{Inquirer} 
{According to each fact-checking scenario $\theta_i$, the Inquirer agent generates the prototype test data: $x \sim q_i(x) = q(x | \theta_i)$, where $q_i(x)$ is the proposal distribution of importance sampling.}
{As depicted in Figure \ref{fig:main_figure}}, a prototype 
{data sample} $x$ encompasses the following {four} components:
\begin{itemize}[leftmargin=*,nosep]
    \item \textit{Key Point} ($\mathcal{KP}$): the specific task instruction for the test case.
    \item \textit{Source Claim} ($\mathcal{SC}$): the claim to be verified.
    \item \textit{Auxiliary Information} ($\mathcal{AI}$): the external knowledge source as the evidences for fact verification.
    \item \textit{Test Mode} ($\mathcal{TM}$): the problem setting of the fact-checking task. Specifically, we consider three widely-studied settings, depending on the type of $\mathcal{AI}$. 1) \textbf{[claim]}: The target LLM verifies $\mathcal{SC}$ without access to external knowledge sources (\textit{i.e.}, $\mathcal{AI}$ remains empty), relying solely on the knowledge stored in its parameters. This setting is widely explored in studies utilizing LLMs for fact-checking~\cite{lee2021towards, wang2024mfc, hu2024large}. 2) \textbf{[evidence]}: $\mathcal{AI}$ is the set of gold evidence from Wiki knowledge that can support or refute the source claim. This setting is also called claim verification~\cite{ma2019sentence}. 3) \textbf{[wisdom of crowds]}: $\mathcal{AI}$ is simulated as the conversation thread on social media towards the source claim. This setting is usually used to verify fake news or rumors by collecting the user interaction as the fact-indicative signal~\cite{shu2019defend, lin2021rumor}.
\end{itemize}

\vspace{-3pt}
\paragraph{Quality Inspector} 
Multiple levels of measures are implemented to guarantee the reliability of the fact-checking questions. To check against the generator role like the Inquirer, we further employ a Quality Inspector agent as the judge role to ensure the diversity of the fact-checking topics and the quality of $\mathcal{AI}$. Especially, in the setting of \textbf{[claim]}, $\mathcal{AI}$ is checked to be empty. In the setting of \textbf{[evidence]}, the Quality Inspector would first integrate external tools to coarsely validate whether the evidence set is more likely from Wiki knowledge via Wikipedia web API, then leverage the rich knowledge embedded in the dominant LLM to finely scrutiny the pieces of evidence. In the setting of \textbf{[wisdom of crowds]}, the Quality Inspector would make sure that the simulated user comments towards the source claim should be valuable enough as the wisdom of crowds for fact verification.

\subsection{Fact Verification with Justification} \label{FVJ}
After obtaining the prototype set of 
{fact-checking test data}, we evaluate target LLMs automatically
{on both fact verification and justification production}. 

\vspace{-3pt}
\paragraph{Evaluator} 
{The Evaluator agent scores the target LLM's predicted verdict and generated justification via LLM-as-a-Judge~\cite{zheng2023judging}, \textit{i.e.}, $\mathcal{F}_\alpha(x)$. In this way, we can assess the fact-checking limits of the target LLM under the hypothesized oracle distribution, \textit{i.e.}, the objective $p(x)\mathcal{F}_\alpha(x)$.}

{Specifically, given a specific test case $x$, the target LLM $\alpha$ generates its response $r$ consisting of predicted verdict and derived justification. Then the output of the Evaluator agent includes an integer rating grade $s\in[1,10]\cap \mathbb{Z}$ and a natural language assessment comment $c$. As a higher score $s$ indicates stronger capabilities, the corresponding fact-checking limitations can be denoted as $\mathcal{F}_\alpha(x) \propto 1/s$.} 
{As illustrated in Figure \ref{fig:main_figure}, we formally define a \textsc{Fact-Audit}'s memory pool as $\mathcal{M}=\{x, r, s,c\}$, which stores the test cases along with their evaluation results. The Evaluator is additionally instructed to distinguish the poorly-performing test cases $\{x|s<\epsilon\}$ based on a predefined threshold $\epsilon$.}
Note that an invalid justification could still get a relatively low grade even if the predicted verdict is correct for fact verification.

\vspace{-3pt}
\paragraph{Prober} 
While collaboration among the four agent roles (\textit{i.e.}, Appraiser, Inquirer, Quality Inspector, and Evaluator) ensures comprehensive fact-checking coverage and model-specific tailoring of our framework, a key challenge lies in effectively identifying areas where the target model underperforms. Although fact-checking prototype test cases provide an intuitive yet superficial assessment of the target LLM's fact-checking capabilities, we argue that they are insufficient to fully reveal the fact-checking limitations and knowledge boundaries due to the inherent constraints of fixed seeds. 

To craft more diverse and unseen test data about each test scenario, we propose iteratively probing for a more comprehensive fact-checking evaluation. Specifically, 
{given the memory pool $\mathcal{M}$ for the current test scenario, the Prober $\mathcal{\rho}$ generates new test data by learning from the model behaviors of the past auditing records stored in $\mathcal{M}$ as the environmental feedback, $x\sim \rho(\mathcal{M})$.}
{Then the Evaluator agent assesses the target LLM on the new test data, and the results are subsequently added to the memory pool.}
Through this iterative probing process, we can effectively identify test data {with poor performance} under each fact-checking test scenario, pinpointing comprehensive insights into the target LLM at the adaptive and different test scenarios.


\begin{table*}[t]
\centering
\scalebox{0.8}{
\begin{tabular}{@{}l|ccccccccc|ccc@{}}
\toprule
\multirow{2}{*}{Model (Target LLM)} & \multicolumn{3}{c}{Complex Claim} & \multicolumn{3}{c}{Fake News} & \multicolumn{3}{c|}{Social Rumor} & \multicolumn{3}{c}{Overall} \\ \cmidrule(lr){2-4} \cmidrule(lr){5-7} \cmidrule(lr){8-10} \cmidrule(lr){11-13}
                       & \textit{IMR}$\downarrow$       & \textit{JFR}$\downarrow$       & \textit{Grade}$\uparrow$     & \textit{IMR}$\downarrow$      & \textit{JFR}$\downarrow$     & \textit{Grade}$\uparrow$   & \textit{IMR}$\downarrow$       & \textit{JFR}$\downarrow$       & \textit{Grade}$\uparrow$     & \textit{IMR}$\downarrow$     & \textit{JFR}$\downarrow$     & \textit{Grade}$\uparrow$   \\ \midrule
\pixtralemoji{}~Mistral-7B             & 60.21     & 25.62          & 3.98      & 47.50    & 19.58        & 4.87    & 59.05     & 39.52          & 3.97      & 54.79   & 23.34        & 4.34    \\ 
\llamaemoji{}~Llama2-7B              & 46.67     & 19.79     & 4.85      & 32.73    & 18.18   & 5.54    & 62.86     & 26.67     & 3.89      & 45.49   & 20.68   & 4.88    \\
\llamaemoji{}~Llama2-13B             & 65.67     & 21.66     & 3.71      & 55.33    & 16.67   & 4.42    & 48.10     & 20.48     & 4.78      & 57.28   & 19.50   & 4.25    \\
\llamaemoji{}~Llama3-8B              & 39.79     & 12.09     & 5.19      & 33.75    & 17.28   & 5.51    & 46.25     & 19.18     & 4.83      & 38.67   & 15.60   & 5.25    \\
\llamaemoji{}~Llama3.1-8B            & 55.83     & 21.46     & 4.36      & 36.39    & 12.78   & 5.60    & 47.62     & 12.86     & 5.00      & 47.52   & 16.77   & 4.91    \\
\llamaemoji{}~Llama3.1-70B           & 41.56     & 14.22     & 5.34      & 25.00    & 11.88   & 6.42    & 38.33     & 10.00     & 5.55      & 34.10   & 12.38   & 5.83    \\
\Qwenemoji{}~Qwen2.5-7B             & 38.97     & 9.74     & 5.38      & 21.54    & 8.20   & 6.58    & 36.67     & 5.42     & 5.68      & 31.76   & 8.14   & 5.91    \\
\Qwenemoji{}~Qwen2.5-72B            & \underline{22.08}     & \underline{5.41}     & \underline{6.62}      & \textbf{10.42}    & \textbf{1.46}   & \textbf{7.67}    & \underline{15.00}     & 3.75     & {7.28}      & \underline{16.00}   & \textbf{3.50}   & \underline{7.17}    \\
\glmemoji{}\hspace{0.13em}GLM4-9B          & 52.73     & 16.36     & 4.76      & 51.67    & 14.00   & 4.93    & 50.00     & 15.24     & 5.00      & 51.67   & 15.24   & 4.88    \\
\Googleemoji{}~Gemma2-9B          & 41.67     & 28.00     & 4.84      & 35.48    & 28.11   & 5.13    & 44.07     & 23.31     & 4.74      & 39.70   & 26.78   & 4.94    \\
\Googleemoji{}~Gemini-Pro             & 30.21     & 11.87     & 5.98      & 19.39    & 5.76   & 6.59    & 32.86     & 5.72     & 5.78      & 27.25   & 8.62   & 6.14    \\
\Claudeemoji{}~Claude3.5-Sonnet      & 32.71     & 9.37     & 6.16      & 15.00    & \underline{2.33}   & \underline{7.41}    & 18.57     & \underline{3.33}     & \underline{7.31}      & 24.34   & {5.96}   & 6.78    \\
\Openaiemoji{}~GPT-4o                 & \textbf{14.05}     & \textbf{4.34}          & \textbf{6.78}      & \underline{10.56}    & 4.93        & 7.26    & \textbf{10.48}     & \textbf{1.41}          & \textbf{7.62}      & \textbf{12.02}   & \underline{3.55}        & \textbf{7.21}    \\
\bottomrule
\end{tabular}}
\vspace{-0.2cm}
\caption{The fact-checking performance of different LLMs audited by \textsc{Fact-Audit}. Metrics include \textit{IMR} (\%), \textit{JFR} (\%), and \textit{Grade}, where \textit{IMR} indicates the insight mastery rate of fact-checking limitations, \textit{JFR} means the flaw rate of the justifications provided by LLMs. The best and second performance are in bold and underlined, respectively.}
\label{tab:main_results}
\vspace{-0.4cm}
\end{table*}

\subsection{
{Adaptive} Updating} \label{IP}

After going through all the existing test scenarios in the fact-checking taxonomy, the Appraiser appeals to new valuable test scenarios, by conducting a critical analysis of instances where the target LLM underperformed in each fact-checking scenario, as indicated by low rating grades in the memory pool $\mathcal{M}$, to unveil potential fact-checking limitations.
Theoretically, the transition probability $\pi(\Theta_{i+1}|\Theta_i,\mathcal{M})$ is estimated, where $\Theta_{i+1}$ is more likely to contain the new test scenario beyond the fact-checking capacities of the target LLM. 
This insight prompts the Appraiser to adaptively refine the taxonomy, ensuring our framework remains relevant and effective in identifying new deficiencies. The cyclical interaction among the Appraiser, Inquirer, and Evaluator establishes a continuous improvement loop, making our auditing framework both comprehensive and responsive to the evolving fact-checking capabilities of different target LLMs.

Finally, after the adaptive updating, the expectation of $\mathcal{F}_\alpha(x)$ in \Cref{eq:obj} for importance sampling can be computed approximately as:
{
\setlength{\abovedisplayskip}{0.1cm}
\setlength{\belowdisplayskip}{0.1cm}
\begin{equation}\small
\begin{aligned}
    &\mathbb{E}_{q(x)}\left[\mathcal{F}_\alpha(x)\frac{p(x)}{q(x)}\right]
        \leq \mathbb{E}_{q(x)}\left[\mathcal{F}_\alpha(x)\right]\\
        &\qquad \qquad \quad \propto \frac{1}{\lvert \mathcal{M}\rvert}\sum_{s\in\mathcal{M}}\frac{1}{s},
\end{aligned}
\end{equation}} 

\noindent where the distributions $q(x)$ and $p(x)$ are intractable in practice. Therefore, since the whole process can only perform limited sampling within the high-probability region ($p(x)/q(x)<1$) of $q(x)$, we compute an upper bound of the target LLM's limitations to effectively reflect its utility.

Overall, this framework enables the adaptive sampling of more targeted and representative fact-checking data, facilitating a comprehensive evaluation of the target LLM's fact-checking capabilities.



\section{Experiments and Results}

\subsection{Experimental Setup}
\vspace{-3pt}
\paragraph{Data} Different from existing static data work, the data within the \textsc{Fact-Audit} agentic framework is dynamically updated to alleviate sampling bias and fairness issues from a fresh perspective. We consider common fact-checking objects such as complex claims, fake news, and social rumors, simulating a diverse real-world data environment.

\vspace{-3pt}
\paragraph{Metric} To audit the fact-checking capacities of LLMs, we introduce three automatic evaluation metrics for quantitative analysis: Insight Mastery Rate (\textit{IMR}), Justification Flaw Rate (\textit{JFR}), and \textit{Grade}. Specifically, \textit{IMR} represents the proportion of low-scoring fact-checking responses relative to the total number of questions, where a \textit{Grade} of three or below (on a ten-point scale) indicates errors in the target LLM’s response, as the Evaluator agent was additionally instructed not to assign a grade higher than three if the target LLM underperformed in either the verdict prediction or justification production stages. \textit{JFR} denotes the percentage of cases where the target LLM conducted correct verdict prediction yet had poor justification, based on the conditions set by \textit{IMR}. \textit{Grade} is assigned by the \textsc{Fact-Audit} framework with employing the scoring prompt inspired by \citet{zheng2023judging}. Overall, \textit{IMR} is the dominant evaluation metric.

\vspace{-3pt}
\paragraph{Target LLMs} To provide a comprehensive LLM auditing, we select 13 representative models as the target LLMs to perform zero-shot inference in \textsc{Fact-Audit}. We adopt ten open-source models: \text{Mistral} (7B)~\cite{jiang2023mistral}, \text{Llama2} (7B, 13B)~\cite{touvron2023llama2}, \text{Llama3} (8B)~\cite{dubey2024llama}, \text{Llama3.1} (8B, 70B), \text{Qwen2.5} (7B, 72B)~\cite{yang2024qwen2}, \text{GLM4} (9B)~\cite{glm2024chatglm}, \text{Gemma2} (9B)~\cite{team2024gemma}; and three proprietary models: \text{Gemini-Pro}~\cite{team2023gemini}, \text{Claude3.5-Sonnet}, \text{GPT-4o}, as our target LLMs. To ensure results are reproducible, the temperature is set as 0 without any sampling mechanism. More implementation details and baseline descriptions are provided in Appendix \S\ref{taxonomy} - \S\ref{EM}.

\subsection{Main Results}
Table~\ref{tab:main_results} presents the auditing results of various LLMs in \textsc{Fact-Audit}, offering a new perspective on fact-checking by incorporating automatic justification production evaluation alongside verdict prediction. Key observations include: 
\begin{itemize}[leftmargin=*,nosep]
    \item \textit{GPT-4o, Qwen2.5-72B, Claude3.5-Sonnet, and Gemini-Pro form the leading tier.} Note that GPT-4o, Claude3.5-Sonnet, and Gemini-Pro are proprietary closed-source models, while Qwen2.5-72B is an open-source model that demonstrates comparable performance in fact-checking evaluation. Besides, GPT-4o achieves the best performance 12.02\% on the dominant metric, \textit{IMR}.
    \item  \textit{The LLaMA series exhibits relatively poorer performance}, spanning the second and third tiers. Llama3-8B and Llama3.1-70B belong to the second tier, alongside Qwen2.5-7B and Gemma2-9B, while other LLaMA models fall into the third tier, showing greater fact-checking limitations on both \textit{IMR} and \textit{Grade} performance.
    \item The auxiliary metric \textit{JFR} of the strong LLM GPT-4o is not the best among all target LLMs, as most low-scoring cases are more likely to be poor justifications when a model excels in factual verdict prediction. This implies \textit{\textsc{Fact-Audit} could elicit the fact-checking limitation of individual target LLMs in accordance with their aptitude}.
    \item LLMs perform relatively well on fake news but struggle with complex claims. This discrepancy may stem from the advanced reasoning capabilities required for complex claims compared to the more factually explicit nature of fake news. The fluctuating performance on social rumors is primarily attributed to their contextual dependence and linguistic complexity, which increase the difficulty of fact-checking for target LLMs. 
\end{itemize}
Overall, the automatic model-centric evaluation, considering justifications beyond verdicts, aligned with intuitive expectations of LLM capabilities and introduced additional fresh dimensions for auditing fact-checking performance and limitations.


\subsection{Analysis of Reliability}

To verify the robustness and fairness of the LLM-generated prototypes, we further conducted the ablative study by adding a setting based on the human-generated prototype seed questions. Specifically, we sampled the same amount of prototypes from the Pinocchio dataset~\cite{hu2024large} as the fixed seed data in \textsc{Fact-Audit}. As shown in Table~\ref{tab:prototype}, it can be observed that the performance of the `LLM-Generated' setting is comparable to that of the `Human-Generated' setting, which highlights the fairness of the LLM auditing in \textsc{Fact-Audit}. We further provided comprehensive human subject studies for quality assurance in Appendix \S\ref{qa} - \S\ref{CTB}. 

\begin{table}[] \large
\centering
\scalebox{0.69}{
\begin{tabular}{@{}l|cccccc@{}}
\toprule
\multirow{2}{*}{Prototype} & \multicolumn{3}{c}{LLM-Generated} & \multicolumn{3}{c}{Human-Generated} \\ \cmidrule(l){2-4} \cmidrule(l){5-7} 
                                  & \textit{IMR}$_{\%}$       & \textit{JFR}$_{\%}$       & \textit{Grade}     & \textit{IMR}$_{\%}$        & \textit{JFR}$_{\%}$        & \textit{Grade}     \\ \midrule
Llama3.1-8B                       & 55.24     & 21.46     & 4.34      & 55.83      & 21.21      & 4.36      \\
Qwen2.5-7B                        & 38.97     & 9.74     & 5.38      & 39.62      & 9.93      & 5.25      \\
GPT-4o                            & 14.05     & 4.34     & 6.78      & 14.24      & 5.02      & 6.59      \\ \bottomrule
\end{tabular}}
\vspace{-0.2cm}
\caption{The comparison of LLM performance based on LLM-generated and human-generated prototypes.}
\label{tab:prototype}
\vspace{-0.1cm}
\end{table}

\subsection{Performance by Test Modes}
To thoroughly examine the impact of different test modes on model performance, we evaluate three representative LLMs (Llama3.1-8B, Qwen2.5-7B, and GPT-4o) in the context of fact-checking. As shown in Table~\ref{tab:test_mode}, we can observe that: 1) [claim] mode is the most challenging for LLMs, as they must rely solely on their parametric knowledge to verify factuality in a closed-book setting. 2) [evidence] mode is the easiest, as all evidence provided is factual and facilitates reasoning, even when conflicting viewpoints are present. 3) [wisdom of crowds] mode falls in the middle. Unlike [claim] mode, it does not depend entirely on the LLM's internal knowledge, and unlike [evidence] mode, it does not explicitly provide guiding signals. Instead, the model must extract valuable insights from the simulated conversation thread to reason effectively. More detailed results are shown in Appendix \S\ref{DPTM}.

\begin{table}[t] \large
\centering
\scalebox{0.7}{
\begin{tabular}{@{}l|c|ccc@{}}
\toprule
{Target LLM}  & {Test Mode} & \textit{IMR}$_{\%}$     & \textit{JFR}$_{\%}$     & \textit{Grade} \\ \midrule
\multirow{3}{*}{Llama3.1-8B} & {[}claim{]}                & 68.80   & 22.87   & 3.56    \\
                             & {[}evidence{]}             & 38.16   & 13.33   & 5.50    \\ 
                             & {[}wisdom of crowds{]}     & 45.29   & 16.08   & 4.96    \\\cmidrule(l){1-2} 
\multirow{3}{*}{Qwen2.5-7B}  & {[}claim{]}                & 48.86   & 12.76   & 4.74    \\
                             & {[}evidence{]}             & 20.83   & 7.31   & 6.45    \\
                             & {[}wisdom of crowds{]}     & 39.58   & 7.40   & 5.43    \\\cmidrule(l){1-2}
\multirow{3}{*}{GPT-4o}      & {[}claim{]}                & 23.05   & 16.67   & 6.11    \\
                             & {[}evidence{]}             & 10.61   & 8.77   & 7.00    \\
                             & {[}wisdom of crowds{]}     & 15.40   & 8.51   & 6.67    \\ \bottomrule
\end{tabular}}
\vspace{-0.2cm}
\caption{The fact-checking performance of three representative LLMs under three fixed test modes.}
\label{tab:test_mode}
\vspace{-0.4cm}
\end{table}

\subsection{Challenging Test Scenarios}
\begin{figure}[t]
    \setlength{\abovecaptionskip}{5pt}   
    \setlength{\belowcaptionskip}{0pt}
  \includegraphics[width=0.47\textwidth]{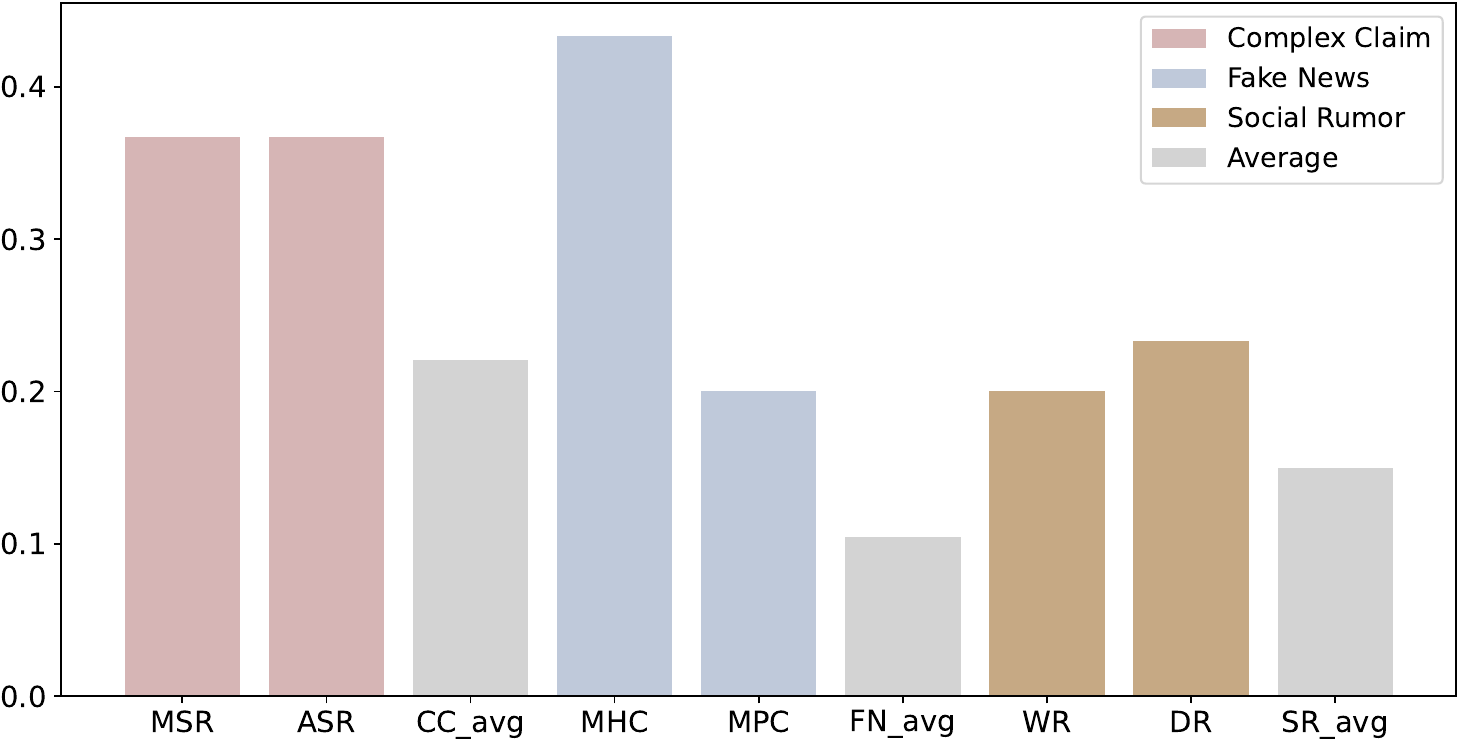}
  \caption{\textit{IMR} of two most challenging test scenarios in each fact-checking objects, with the averaged \textit{IMR}.}
  \label{fig:poor_scenario}
  \vspace{-0.2cm}
\end{figure}

As shown in Figure~\ref{fig:poor_scenario}, we conduct an analysis to discuss the challenging test scenarios in \textsc{Fact-Audit}, by taking the \textit{IMR} performance of the well-performed open-source LLM Qwen2.5-72B as an example. We can see that: (1) ``Multi-Step Reasoning'' (MSR) and ``Aggregated Statistical Reasoning'' (ASR),  (2) News content with ``Mismatched Headline or Caption'' (MHC) and ``ManiPulated Content'' (MPC), and (3) ``Wishing Rumor'' (WR) and ``Dreading Rumor'' (DR), are the two most challenging scenarios for Complex Claim, Fake News, and Social Rumor, respectively. Besides, although the Fake News averaged (FN\_avg) \textit{IMR} is lower than the Complex Claim averaged (CC\_avg) and Social Rumor averaged (SR\_avg), the detailed MHC scenario of Fake News is the most difficult than those scenarios of Complex Claim and Social Rumor. We additionally provided more discussion about fact-checking topics in Appendix \S\ref{DFT} - \S\ref{DAU}.

\subsection{Iterative Probing}
\begin{figure}[t]
    \setlength{\abovecaptionskip}{5pt}   
    \setlength{\belowcaptionskip}{0pt}
  \includegraphics[width=0.47\textwidth]{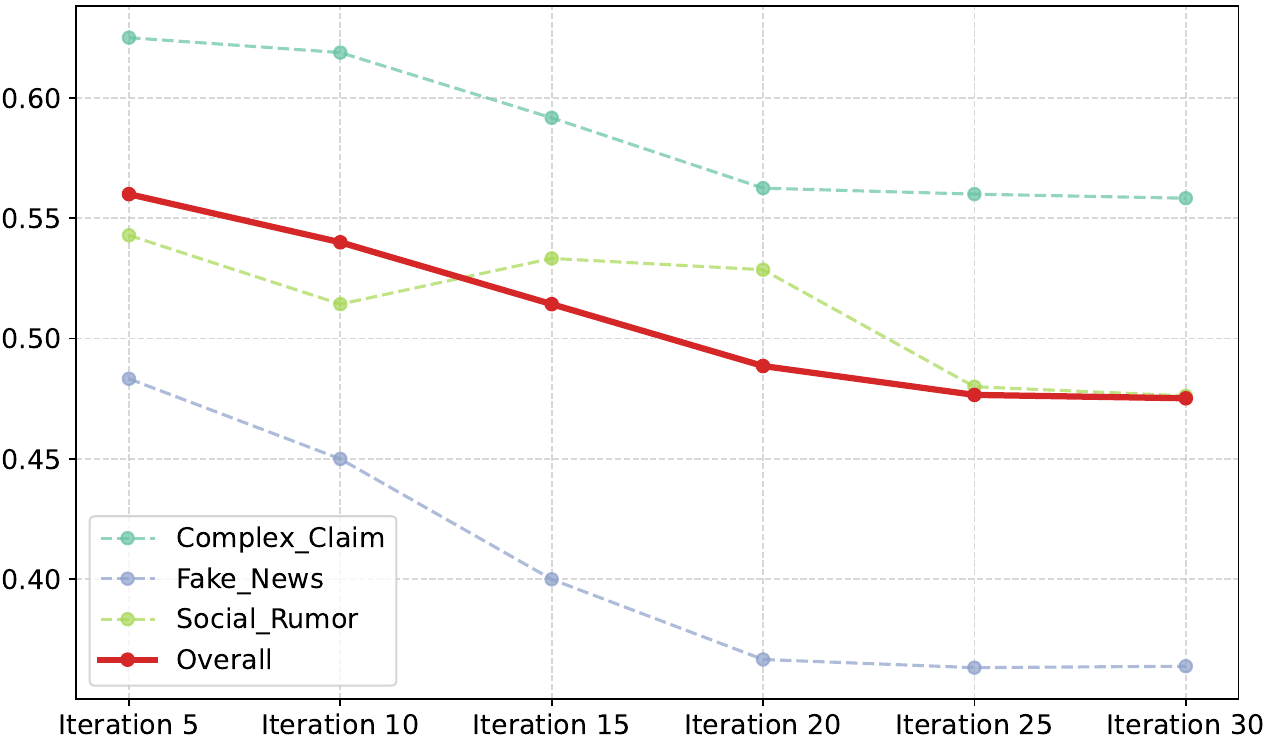}
  \caption{The effect of different iterations in scaling.}
  \label{fig:eip}
  \vspace{-0.3cm}
\end{figure}

We analyze the effect of multi-turn iterative probing in \textsc{Fact-Audit} by examining the \textit{IMR} performance across different iterations, as illustrated in Figure~\ref{fig:eip}. The results show that the \textit{IMR} metric decreases as the number of iterations increases, eventually converging. As the test data expands, the model's performance is more comprehensively evaluated, allowing the identification of truly problematic cases that represent the model's inherent weaknesses. This approach enables a deeper exploration of less obvious model limitations while reinforcing the flexibility of our evaluation framework to scale seamlessly with the size of the assessment.

\subsection{Case Study} \label{case_study}

\begin{figure*}
    \centering
    \includegraphics[width=1\linewidth]{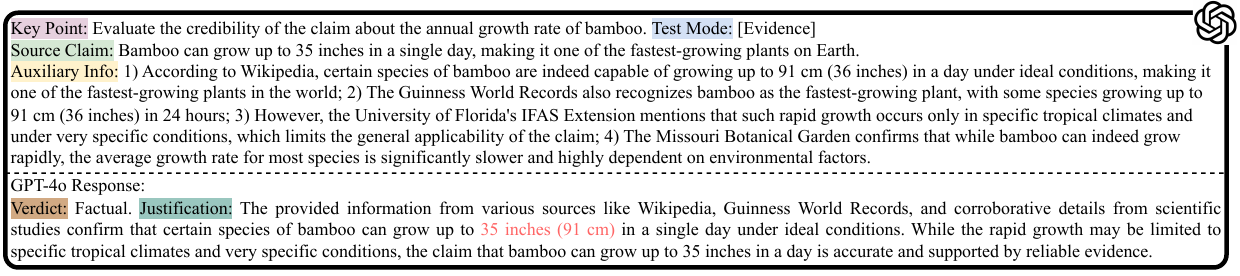}
    \vspace{-0.7cm}
    \caption{Example of cases with correct verdict prediction yet poor justification for fact-checking the claim.}
    \label{fig:case}
    \vspace{-0.3cm}
\end{figure*}
One key advantage of \textsc{Fact-Audit} is that the limitations of fact-checking in LLMs where poor justification is provided for the correct verification could be well probed. To better understand how \textsc{Fact-Audit} proactively unveil such limitations, we conduct a case study on the GPT-4o's response to the test data sample, as exemplified in Figure~\ref{fig:case}. We can observe that there is a factual error in the provided justification. The target LLM states that bamboo can grow ``up to 35 inches (91 cm)'', which conflicts with the unit conversion knowledge that 35 inches is equivalent to 88.9 cm. Even though the related knowledge is provided in the auxiliary information, the target LLM still failed to provide precise justification for fact-checking the claim. This reaffirms the importance of incorporating justification into automatic evaluations, enabling more comprehensive auditing beyond merely assessing accuracy. More cases are shown in Appendix \S\ref{more_case}.

\section{Related Work}
\vspace{-3pt}
\paragraph{Fact-Checking Evaluation} 
Automated fact-checking has gained significant attention in the NLP research community in recent years as a means of combating misinformation and disinformation. Various datasets have been proposed that enable the development and evaluation of systems for automatic fact-checking, the most popular ones being based on human-crafted claims from Wikipedia content~\cite{thorne2018fever, sathe2020automated, schuster2021get}, claims in fake news published by a news outlet~\cite{buntain2017automatically, shu2020fakenewsnet, nakov2022clef}, rumorous claims on social media~\cite{ma2015detect, ma2017detect, lin2022detect}, complex claims that require multi-step reasoning~\cite{jiang2020hover, aly2021feverous}, naturally occurring claims in specific domains~\cite{gupta2021x, wadden2022scifact, lin2023zero}, and LLM-generated misinformation~\cite{chen2024can}, etc. To understand the factual knowledge of LLMs, \citet{hu2024large} curated a new fact-checking benchmark by organizing previous representative datasets, aiming to identify weaknesses in LLM fact verification. However, besides the inevitable issue of test set leakage, this static evaluation approach relied primarily on expert-designed, specialized tasks from existing datasets, overlooking emerging LLM-generated content and lacking adaptability to the complex, open-ended nature of real-world applications. Different from previous work on static accuracy evaluation, leveraging the derived justification~\cite{atanasova2020generating, guo2022survey} from LLMs, our work aims to explore the dynamic auditing beyond the veracity prediction, to dynamically elicit the limitations of fact-checking in LLMs.

\vspace{-3pt}
\paragraph{LLM Agent} 
The integration of LLMs as agents spans various domains, such as code generation and game-playing, demonstrating their robust planning and reasoning capabilities across diverse contexts~\cite{wang2023voyager, yao2022react, shen2023hugginggpt, mu2023embodiedgpt, hong2023metagpt, liu2023agentbench, sun2023adaplanner, qian2023communicative}. These advancements highlight the ability of LLMs to handle complex tasks with minimal supervision. In parallel, self-improvement methodologies~\cite{chen2022codet, chen2023teaching, shinn2023reflexion, madaan2023self} have emerged, utilizing feedback-driven processes to iteratively enhance output quality. Building on these insights, we develop a novel agentic framework for systematical LLM auditing in fact-checking complex claims, fake news or rumors.

\section{Conclusion and Future Work}
We introduced \textsc{Fact-Audit}, an adaptive multi-agent evaluation framework that dynamically elicits the fact-checking limitations of LLMs. By automatically evaluating the justification production beyond the verdict prediction, \textsc{Fact-Audit} enables scalable, model-centric LLM auditing for fact-checking tasks. Experiments on a dozen mainstream LLMs reveal a notable performance gap between closed and open-source models with different sizes. In future work, we plan to further exploit the reliability of the proposed framework.

\section*{Limitations}
There are multiple ways for further improvement of this work to alleviate the following limitations:
\begin{itemize}
    \item Firstly, despite implementing various measures, such as error-correction mechanisms and human evaluations, to enhance the stability and transparency of the agent controller and reduce bias and errors, we argue that the potential biases in fact-checking (much like those inherent to humans ) remain unavoidable. Even human beings or most advanced models have knowledge bias. In future research, we will continue updating the evaluation framework to a more robust and reliable evaluation framework. This would constitute another targeted area of research.
    \item Secondly, despite its vast knowledge reserves, the agent controller is constrained by its limited ability to acquire and integrate new information dynamically. This limitation hinders its capacity to adapt to evolving knowledge landscapes. In future work, we aim to incorporate advanced techniques such as Retrieval-Augmented Generation (RAG) to enhance the agent's decision-making capabilities, enabling it to access up-to-date information and provide more accurate, context-aware responses.
    \item Lastly, while our multi-agent evaluation framework adaptively and dynamically identifies specific deficiencies in target LLMs related to fact verification and justification production, it currently lacks an effective mechanism for model improvement. In future work, we aim to integrate preference optimization methodologies, enabling the framework not only to audit the fact-checking capabilities of LLMs and generate actionable insights for performance refinement but also to provide high-quality training data to facilitate effective model improvement.
\end{itemize}

\section*{Ethics Statement}
This research involved human subject studies to evaluate the quality and reliability of \textsc{Fact-Audit}. The following considerations were adhered to ensure the protection and ethical treatment of participants: 1) Voluntary Participation: All participants were informed about the nature of the research and their role in it. Participation was entirely voluntary, with participants having the right to withdraw at any time without any consequences. 2) Informed Consent: Written informed consent was obtained from all participants. This consent form detailed the purpose of the research, the procedures involved, potential risks, and measures taken to safeguard participant data. 3) Data Anonymity and Confidentiality: All data collected during the study were anonymized. Personal identifiers were removed to maintain confidentiality and data were stored securely to prevent unauthorized access. 4) Minimal Risk: The study involved minimal risk to participants. The tasks performed were similar to everyday activities, and no sensitive personal information was requested or recorded.

Research indicates that evaluating content related to misinformation can have negative effects. To protect our human evaluators, we establish three guidelines: 1) ensuring their acknowledgment of viewing potentially misleading content, 2) limiting weekly evaluations and encouraging a lighter daily workload, and 3) advising them to stop if they feel overwhelmed. Finally, we regularly check in with evaluators to ensure their well-being.

The purpose of this work is to prevent the spread of misinformation/disinformation and to ensure that people are not subjected to non-factual information. Nevertheless, we are aware of the potential for malicious users to reverse-engineer and create misinformation guided by \textsc{Fact-Audit}. This is strongly discouraged and condemned. Furthermore, all the fact-checking test data generated by the agents do not contain any personal information.



\bibliography{custom}

\appendix



\section{Taxonomy}\label{taxonomy}

\begin{figure}[t]
  \includegraphics[width=\columnwidth]{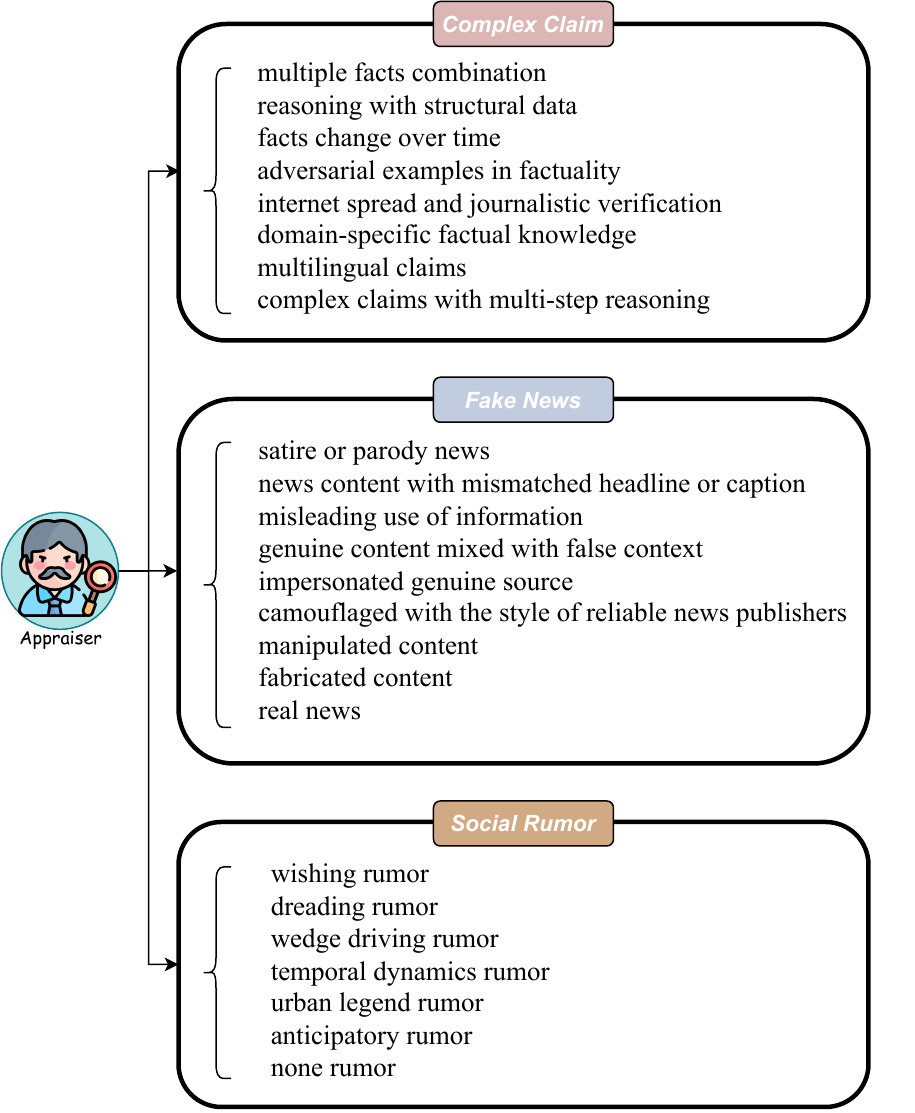}
  \vspace{-0.7cm}
  \caption{Illustration of the initial taxonomy on the fact-checking objects: Complex Claim, Fake News, Social Rumor.}
  \label{fig:categorization}
  \vspace{-0.3cm}
\end{figure}

We provide the initial taxonomy in Figure~\ref{fig:categorization}. The detailed taxonomy of the three fact-checking objects draws the practice of the previous fact-checking literature: 1) Complex Claim~\cite{pan2023fact, hu2024large} involves assertions that require detailed analysis and support from multiple sources and are common in scientific or academic discussions, 2) Fake News~\cite{waldrop2017genuine} refers to deliberately fabricated or distorted information aimed at misleading audiences, often seen on social media to influence public opinion or for economic gain, and 3) Social Rumor~\cite{allport1947psychology} is a piece of information that spreads quickly and remains to be verified, usually through word of mouth or social media, and can lead to misunderstandings or unnecessary panic. The choice of complex claims, fake news, and rumors as fact-checking objects stems from their prominent impact on public discourse and their prevalence in today's information landscape. In \textsc{Fact-Audit}, the final test scenarios of each fact-checking object would be evolved and updated according to the model-specific performance. Due to the dynamic nature, we provide the averaged statistics of the data in \textsc{Fact-Audit} as shown in Table~\ref{tab:statistic}.

\begin{table*}[t]
\centering
\scalebox{0.8}{
\begin{tabular}{@{}l|ccc|c@{}}
\toprule
Model            & Complex Claim & Fake News & Social Rumor & Overall \\ \midrule
Mistral-7B       & 480           & 480       & 210          & 1170    \\
Llama2-7B        & 480           & 330       & 210          & 1020    \\
Llama2-13B       & 300           & 300       & 210          & 810     \\
Llama3-8B        & 480           & 480       & 240          & 1200    \\
Llama3.1-8B      & 480           & 360       & 210          & 1050    \\
Llama3.1-70B     & 450           & 480       & 240          & 1170    \\
Qwen2.5-7B       & 390           & 390       & 240          & 1020    \\
Qwen2.5-72B      & 480           & 480       & 240          & 1200    \\
GLM4-9B          & 330           & 300       & 210          & 840     \\
Gemma2-9B        & 300           & 420       & 270          & 990     \\
Gemini-Pro       & 480           & 330       & 210          & 1020    \\ 
Claude3.5-Sonnet & 480           & 300       & 210          & 990     \\
GPT-4o & 420           & 360       & 210          & 990     \\\bottomrule
\end{tabular}}
\caption{The averaged data statistics of the dynamically-updated auditing framework corresponding to each specific target LLM.}
\label{tab:statistic}
\end{table*}

The taxonomy of fact-checking objects is systematically designed to address the diverse forms of misinformation based on their intrinsic characteristics, verification challenges, and real-world impact. Specifically, the following principles are used to guide the categorization: 1) Complexity: The level of reasoning and factual knowledge required to validate the claim. 2) Intent and Structure: Whether the content aims to mislead, parody, or inform and how it is presented. 3) Propagation Dynamics: The nature and speed at which rumors or misinformation spread within social contexts. The proposed taxonomy serves as the foundation for systematically evaluating fact-checking capacities in LLMs. By dividing fact-checking objects into Complex Claims, Fake News, and Social Rumors, the framework achieves the following objectives: 1) Targeted Evaluation: Addressing the unique verification challenges posed by each category. 2) Comprehensive Coverage: Ensuring that the taxonomy encompasses a wide range of misinformation types prevalent in real-world scenarios. 3) Practical Utility: Facilitating the generation of more targeted and representative fact-checking datasets to evaluate model performance. This taxonomy is designed to systematically uncover fact-checking limitations in LLMs by segmenting diverse fact-checking objects into detailed test scenarios. Each category reflects the nature, complexity, and propagation style of the potential true or false information, enabling a more structured and comprehensive evaluation framework.


\section{Implementation Details}
\label{implementation_details}
For all experiments, we adopt GPT-4o as the core model for \textsc{Fact-Audit}. For importance sampling, we formalize the probability density of each fact-checking data as a uniform distribution, to mitigate potential long-tail issues. Compared results ($p < 0.05$ under t-test) are averaged over three random runs. The maximum number of iterations is set to 30 for evaluations on each fact-checking test scenario. The threshold $\epsilon$ for the poorly-performing test cases is set as 4.0. The cost for evaluating one target LLM is about 25 dollars and 6 hours. All experiments were conducted using two NVIDIA A100 80GiB GPUs.
In the following, the details of agent implementation would be depicted. 

\begin{figure}[]
  \includegraphics[width=\columnwidth]{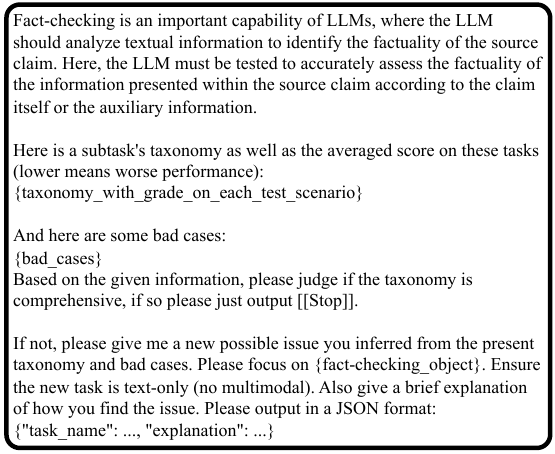}
  \vspace{-0.7cm}
  \caption{Instruction for the Appraiser agent to generate new valuable test scenarios.}
  \label{fig:prompt1}
  \vspace{-0.3cm}
\end{figure}

\textbf{Appraiser.} For the Appraiser agent, the taxonomy is initialized as shown in Figure~\ref{fig:categorization}. We set the temperature of the Appraiser agent as the default setting of 1.0. To update the taxonomy, the instruction prompt is used as shown in Figure~\ref{fig:prompt1}. If the Appraiser outputs the ``[stop]'' tokens in three times, the adaptive updating process would be terminated. 

\begin{figure}[]
  \includegraphics[width=\columnwidth]{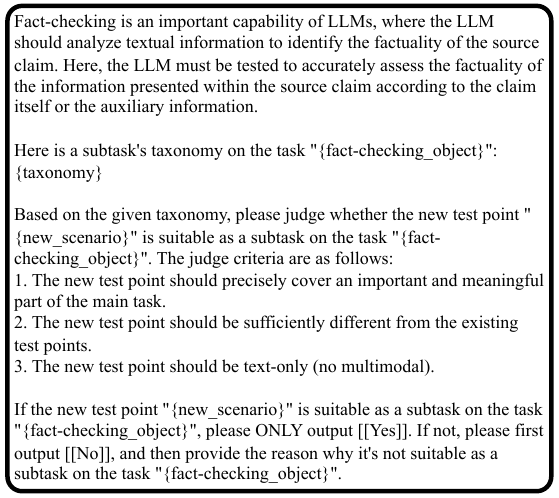}
  \vspace{-0.7cm}
  \caption{Instruction for the Appraiser agent to check whether the new proposed test scenario is suitable to be added into the current taxonomy.}
  \label{fig:prompt2}
  \vspace{-0.3cm}
\end{figure}

Note that as the new subject task cannot be always added, we additionally apply a judge agent to check the quality of the new proposed test scenario. The prompt for judging the new test scenario is shown in Figure~\ref{fig:prompt2}.

\begin{figure}[]
  \includegraphics[width=\columnwidth]{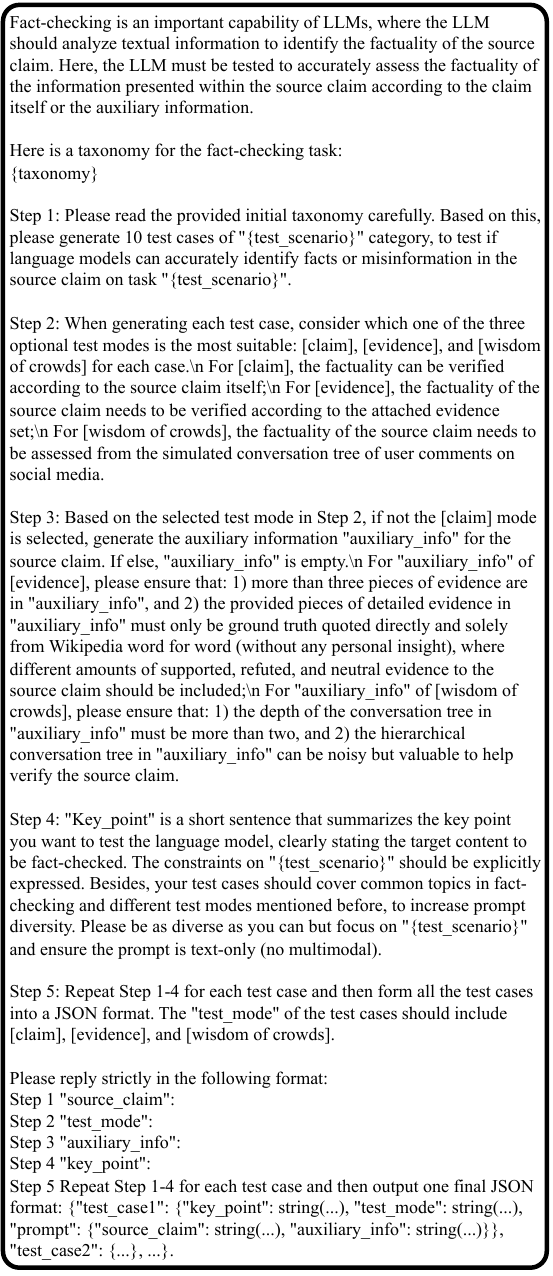}
  \vspace{-0.7cm}
  \caption{Instruction for the Inquirer agent to generate the prototype fact-checking data.}
  \label{fig:prompt3}
  \vspace{-0.3cm}
\end{figure}

\textbf{Inquirer.} The role of the Inquirer is to generate the prototype fact-checking data. To ensure the fairness, we set the temperature of the Inquirer agent as 0.0 without any sampling mechanism. The number of prototype seed questions for each test scenario is set as 10. The instruction prompt is designed as shown in Figure~\ref{fig:prompt3}.

\begin{figure}[]
  \includegraphics[width=\columnwidth]{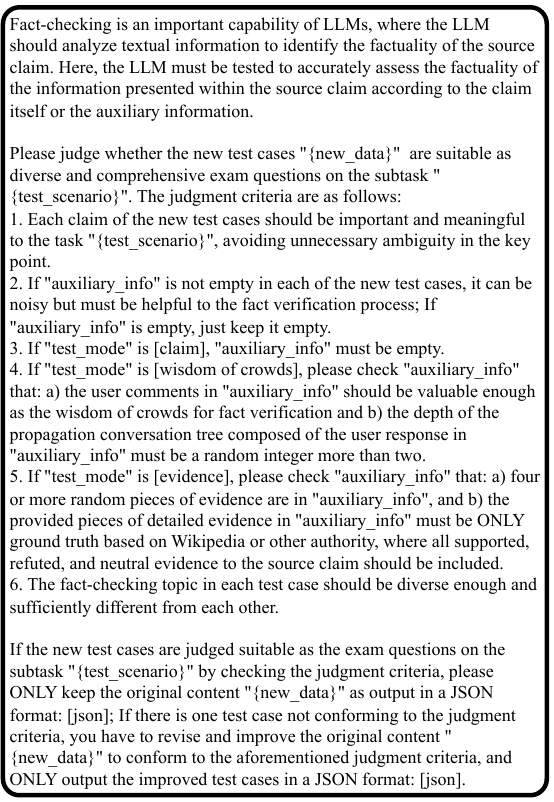}
  \vspace{-0.7cm}
  \caption{Instruction for the Quality Inspector agent to improve the quality of the fact-checking data.}
  \label{fig:prompt4}
  \vspace{-0.3cm}
\end{figure}

\textbf{Quality Inspector.} To ensure the quality of the LLM-generated fact-checking data, the Quality Inspector agent is deployed to use external tools and inspect whether the generated data conforms to the basic requirements. The parameter temperature is set as 0.0 since here we do need the most reliable content instead of the generation model's creativity. First of all, the Wikipedia search API is called to coarsely check the credibility of the auxiliary information if the test scenario is under the [evidence] mode. Then the specific prompt is curated finely as shown in Figure~\ref{fig:prompt4}. 

\begin{figure}[]
  \includegraphics[width=\columnwidth]{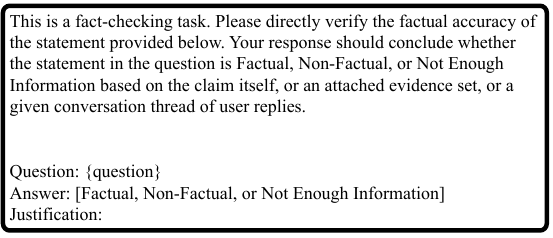}
  \vspace{-0.7cm}
  \caption{Instruction for generating the reference answer of the Evaluator agent and the answer of the target LLM for fact-checking data.}
  \label{fig:prompt_}
  \vspace{-0.3cm}
\end{figure}

\begin{figure}[]
  \includegraphics[width=\columnwidth]{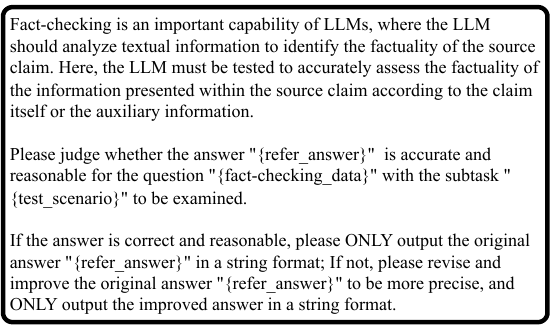}
  \vspace{-0.7cm}
  \caption{Instruction for the Evaluator agent to ensure the quality of the reference answers for fact-checking data.}
  \label{fig:prompt5}
  \vspace{-0.3cm}
\end{figure}

\textbf{Evaluator.} Some recent pioneering benchmark work~\cite{zheng2023judging, dubois2024length, cheng2024autodetect} used LLM-as-a-judge to ask the strong LLM to compare model responses to a static dataset of questions. The prompt template of the question $\{\mathcal{SC}, \mathcal{AI}\}$ for the reference answer of the Evaluator and the answer of the target LLM is shown in Figure~\ref{fig:prompt_}. The model's judgments achieved over 80\% agreement with human preferences, proving the usability of using LLMs to evaluate response quality. Inspired by the previous literature, we employ an Evaluator agent to evaluate the response of the target LLM in a scoring and comparison-based manner. First, we employ three agent controllers (temperature = 1.0) with the currently dominant LLM GPT-4o, to vote a relatively perfect answer in a self-reflection manner. Nevertheless, to further ensure the quality of the reference answer, we made another judgment agent role (temperature = 0.0) to check the content of the reference answer generated by the Evaluator agent, where the prompt is shown in Figure~\ref{fig:prompt5}. 

\begin{figure}[]
  \includegraphics[width=\columnwidth]{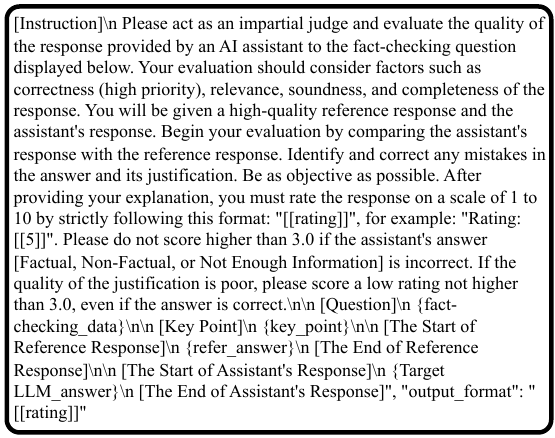}
  \vspace{-0.7cm}
  \caption{Instruction for the Evaluator agent to conduct the fact-checking evaluation in an LLM-as-a-Judge manner.}
  \label{fig:prompt6}
  \vspace{-0.3cm}
\end{figure}

In this way, we can alleviate the potential mistakes in the reference answer. Then, we employ the scoring prompt~\cite{zheng2023judging}, to elicit an evaluation output consisting of a grade and a comment on the response of the target LLM. Specifically, the prompt is devised as shown in Figure~\ref{fig:prompt6}.

Note that in our designed evaluation setup, a \textit{Grade} of three or below (on a ten-point scale) was selected to represent errors in the target LLM’s response, based on our analysis of quality differentiation and the practice of previous literature~\cite{cheng2024autodetect}. This threshold effectively captures significant issues in either the verdict prediction or justification production stages, ensuring that only responses demonstrating an adequately reliable quality receive a higher score. As we employ the ten-point scale in the evaluation, considering different levels of uniform division, a grade under 4.0 naturally represents a low level, while 7.0 is the dividing line between medium and high levels, which is a reasonable setting in our evaluation system. This threshold was determined through careful consideration of maintaining strict reliability and consistency across the subsequent introduced evaluation metrics.

\begin{figure}[]
  \includegraphics[width=\columnwidth]{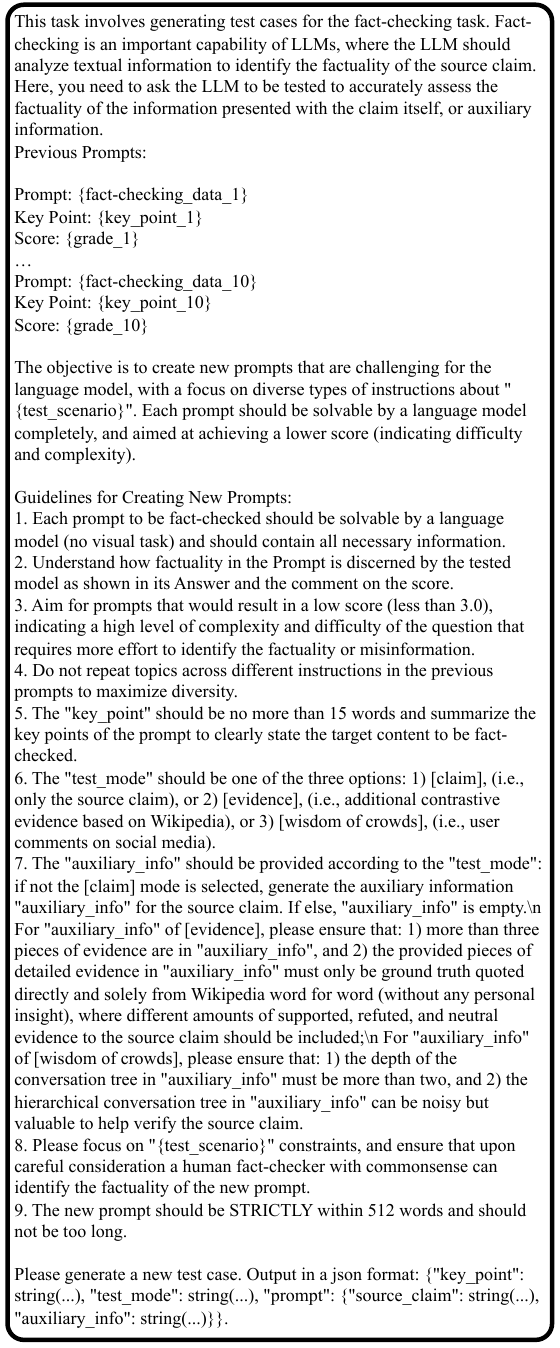}
  \vspace{-0.7cm}
  \caption{Instruction for the Prober agent to generate more diverse and unseen fact-checking data.}
  \label{fig:prompt7}
  \vspace{-0.3cm}
\end{figure}

\textbf{Prober.} All the evaluation output and fact-checking data would be recorded in a memory pool. Based on the collection of the evaluation history in the memory pool, we deploy a Prober agent (temperature = 1.0) to further explore more comprehensive fact-checking data to query the target LLM. The concrete instruction prompt is designed as shown in Figure~\ref{fig:prompt7}.

For the baselines, we conduct extensive experiments in \textsc{Fact-Audit} to evaluate a total of 13 representative target LLMs:
\begin{itemize}
    \item \text{Mistral-7B}: A highly efficient 7-billion parameter open-source large language model optimized for performance, offering state-of-the-art results in various natural language processing tasks while maintaining lightweight computational requirements. We specifically utilize the ``Mistral-7B-Instruct-v0.2'' version.
    \item \text{Llama2-7B}: An advanced 7-billion parameter open-source large language model developed by Meta, designed to deliver strong performance in natural language understanding and generation tasks, with fine-tuning options for specialized applications. We specifically utilize the ``Llama-2-7b-hf'' version.
    \item \text{Llama2-13B}: A 13-billion version of LLaMA 2 series. We specifically utilize the ``Llama-2-13b-hf'' version.
    \item \text{Llama3-8B}: An 8-billion parameter large language model released by Meta in April 2024 as part of the LLaMA 3 series, optimized for dialogue and conversational tasks with the ability to generate natural language text. We specifically utilize the ``Meta-Llama-3-8B-Instruct'' version.
    \item \text{Llama3.1-8B}: An 8-billion parameter large language model released by Meta in July 2024, supporting multilingual dialogue with a 128,000-token context window, designed for various natural language processing tasks. We specifically utilize the ``Meta-Llama-3.1-8B-Instruct'' version.
    \item \text{Llama3.1-70B}: A 70-billion version of LLaMA 3.1 series. We specifically utilize the ``Meta-Llama-3.1-70B-Instruct'' version.
    \item \text{Qwen2.5-7B}: A 7-billion parameter large language model from the Qwen2.5 series, offering enhanced capabilities in coding, mathematics, and instruction following, with support for over 29 languages and a context length of up to 131,072 tokens. We specifically utilize the ``Qwen2.5-7B-Instruct'' version.
    \item \text{Qwen2.5-72B}: A 72-billion version of Qwen 2.5 series. We specifically utilize the ``Qwen2.5-72B-Instruct'' version.
    \item \text{GLM4-9B}: An open-source large language model developed by Zhipu AI, featuring 9 billion parameters and supporting 26 languages, including Japanese, Korean, and German. We specifically utilize the ``glm-4-9b-chat'' version.
    \item \text{Gemma2-9B}: An open-source language model developed by Google, trained on 8 trillion tokens of diverse data, including web documents, code, and mathematical texts, to deliver state-of-the-art performance across various natural language processing tasks. We specifically utilize the ``gemma-2-9b-it'' version.
    \item \text{Gemini-Pro}: A multimodal large language model developed by Google DeepMind, designed to process and generate text, audio, and visual data, offering advanced reasoning and planning capabilities across various tasks. We specifically utilize the ``Gemini-Pro'' version.
    \item \text{Claude3.5-Sonnet}: Anthropic's most advanced large language model, offering enhanced reasoning, coding proficiency, and vision capabilities, with a 200,000-token context window and improved speed and cost efficiency. We specifically utilize the ``claude-3-5-sonnet-20240620'' version.
    \item \text{GPT-4o}: The latest flagship model developed by OpenAI, designed for real-time reasoning across audio, visual, and textual inputs. We specifically utilize the ``gpt-4o-2024-05-13'' version.
\end{itemize}
To ensure the reproducibility, we set the temperature as 0.0 without any sampling mechanism.

\section{Evaluation Metrics} \label{EM}
To audit the fact-checking capacities of LLMs, we introduce three automatic evaluation metrics for quantitative analysis: Insight Mastery Rate (\textit{IMR}), Justification Flaw Rate (\textit{JFR}), and \textit{Grade}. Specifically, \textit{IMR} represents the proportion of low-scoring fact-checking responses relative to the total number of questions, where a \textit{Grade} of three or below (on a ten-point scale) indicates errors in the target LLM’s response, as the Evaluator agent was additionally instructed not to assign a grade higher than three if the target LLM underperformed in either the verdict prediction or justification production stages. 
Specifically, the \textit{IMR} metric can be formulated as follows:
\begin{equation}\small
    \textit{IMR} = \frac{\text{Number of Tests with Grade $\leq$ 3.0}}{\text{Total Number of Tests}},
\end{equation} where \textit{IMR} represents the degree of mastering the fact-checking limitation insight of target LLMs.

\textit{JFR} denotes the percentage of cases where the target LLM conducted correct verdict prediction yet had poor justification, based on the conditions set by \textit{IMR}. Specifically, the \textit{JFR} metric can be formulated as follows:
\begin{equation}\small
    \textit{JFR} = \frac{\text{Number of Tests with CVPJ}}{\text{Total Number of Tests}},
\end{equation} where CVPJ denotes the case that the target LLM predicted Correct Verdict, but provided a relatively Poor Justification.

\textit{Grade} is assigned by the \textsc{Fact-Audit} framework with employing the scoring prompt inspired by \citet{zheng2023judging}. Overall, \textit{IMR} is the dominant evaluation metric.

\section{Quality Assurance}\label{qa}
To guarantee the reliability of the fact-checking data generated by \textsc{Fact-Audit} agents, 3 professional fact-checking annotators (between the ages of 26 and 29) were asked to judge whether the quality of fact-checking data on each sample was up to standard or not, including the source claim, the auxiliary information in the test mode [evidence] or [wisdom of crowds], and the reference answer. Thus we randomly sampled 600 pieces, with 200 from each fact-checking object, across all evaluations of target LLMs. 

Specifically, as shown in Table~\ref{tab:quality}, the annotators, with the averaged intra-class agreement score 0.669, need to evaluate: 1) whether the factual knowledge taxonomy in the categorization is suitable as the test scenarios in the fact-checking task, for the quality judgment of the detailed categorization; 2) whether the claim is check-worthy in the fact-checking process, for the quality judgment of the source claim; 3) whether the supported, refuted, and neutral snippets included in the auxiliary information to the source claim are all ground truth, for the quality judgment of the relevant evidence as auxiliary information; 4) whether the simulated conversation thread is valuable to conduct fact verification, for the quality judgment of the wisdom of crowds as auxiliary information; 5) whether the reference answer is reasonable and correct, for the quality judgment of the reference answer; 6) whether the grade and comment scored by the agent are acceptable for the auditing basis, for the quality judgment of the output evaluation.

\begin{table}[t]
\centering
\scalebox{0.8}{
\begin{tabular}{@{}lcc@{}}
\toprule
Fact-checking Data            & Judgment$\uparrow$ & Agreement$\uparrow$ \\ \midrule
Detailed Taxonomy    & 98.86   & 0.810       \\
Source Claim   & 97.17    & 0.795     \\
Relevant Evidence        & 87.00    & 0.619     \\
Wisdom of Crowds & 81.83    & 0.581     \\
Reference Answer & 90.33    & 0.762     \\
Output Evaluation & 89.02    & 0.658 \\ \bottomrule
\end{tabular}}
\caption{Human subject study on the reliability of the \textsc{Fact-Audit} framework. The Judgment (\%) means the proportion of fact-checking data that meets the criteria, and the Agreement denotes the average Cohen's Kappa between any two expert annotators.}
\label{tab:quality}

\end{table}

We can observe that: 1) The highest judgment indicates that the taxonomy used for fact-checking tasks is highly suitable, demonstrating the reliability of Fact-Audit in designing test scenarios. 2) Most source claims were judged as check-worthy, reflecting the high quality of the claims generated for fact-checking tasks. 3) For the provided relevant evidence, while still high, the quality score here is slightly lower. This may be due to the complexity of the supporting, refuting, or neutral snippets provided as auxiliary evidence. 4) For the simulated conversation thread as wisdom of crowds, the lowest judgment (81.83\%) suggests that extracting valuable information from simulated conversations poses significant challenges. The unstructured nature or semantic ambiguity in dialogues may contribute to this difficulty. 5) Over 90\% of the reference answers being correct reflects their good quality. 6) Furthermore, the acceptability of the Evaluator agent's grades and comments is robust, with a high-quality judgment (89.02\%) indicating reliability in the auditing process. 

Detailed Taxonomy (0.810) and Source Claim (0.795) demonstrate very high agreement, surpassing or approaching the 0.8 threshold, indicating the objectivity and reliability of these evaluations. Relevant Evidence (0.619) and Wisdom of Crowds (0.581) have lower agreement, especially Wisdom of Crowds (0.581), which falls within the moderate range. This suggests that these tasks involve more subjective judgments and are more challenging to evaluate consistently. Reference Answer (0.762) and Output Evaluation (0.658) show reasonable agreement, although slightly less consistent compared to the detailed
taxonomy and source claims.

\section{Comparison with Traditional Benchmarks} \label{CTB}




\begin{table}[t]
\centering
\scalebox{0.8}{
\begin{tabular}{@{}lccc@{}}
\toprule
Benchmark   & Pinocchio & \textsf{LLMFake} & \textbf{\textsc{Fact-Audit}} \\ \midrule
Redundancy$\downarrow$  & 2.03      & 2.31    & 1.22       \\
Diversity$\uparrow$   & 1.94      & 2.17    & 2.62       \\
Readability$\uparrow$ & 2.86      & 2.43    & 2.91       \\
Coverage$\uparrow$    & 2.14      & 1.65    & 2.58       \\
Fairness$\uparrow$    & 2.57      & 2.53    & 2.56       \\
Suitability$\uparrow$ & 2.79      & 2.78    & 2.81       \\ \bottomrule
\end{tabular}}
\vspace{-0.2cm}
\caption{Human evaluation of the benchmark quality.}
\vspace{-0.4cm}
\label{tab:benchmark_comparison}
\end{table}

We conduct the human subject study to compare the benchmark quality of our proposed \textsc{Fact-Audit} and other two well-known benchmarks Pinocchio~\cite{hu2024large} and \textsf{LLMFake}~\cite{chen2024can} used for automated LLM fact-checking evaluation. We randomly selected 450 samples, with 150 from each benchmark. Three professional fact-checking annotators (between the ages of 26-29) were asked to evaluate the data quality according to the following criteria: 1) Redundancy: the repetitiveness or unnecessary duplication within the data; 2) Diversity: the variety and range of different test scenarios set by the data; 3) Readability: how easy it is for humans to read and understand the content; 4) Coverage: how comprehensively the dataset covers the relevant subjects or topics; 5) Fairness: whether the data presents information in a balanced and unbiased manner; 6) Suitability: the appropriateness of the data for automatic fact-checking evaluation. For each criterion, a 3-point Likert scale was employed, where 1 meant the poorest quality and 3 the best. 

\begin{figure*}[ht]
\centering
\subfigure[Pinocchio]{
\begin{minipage}[t]{0.33\linewidth}
\centering
\scalebox{0.85}{\includegraphics[width=6cm]{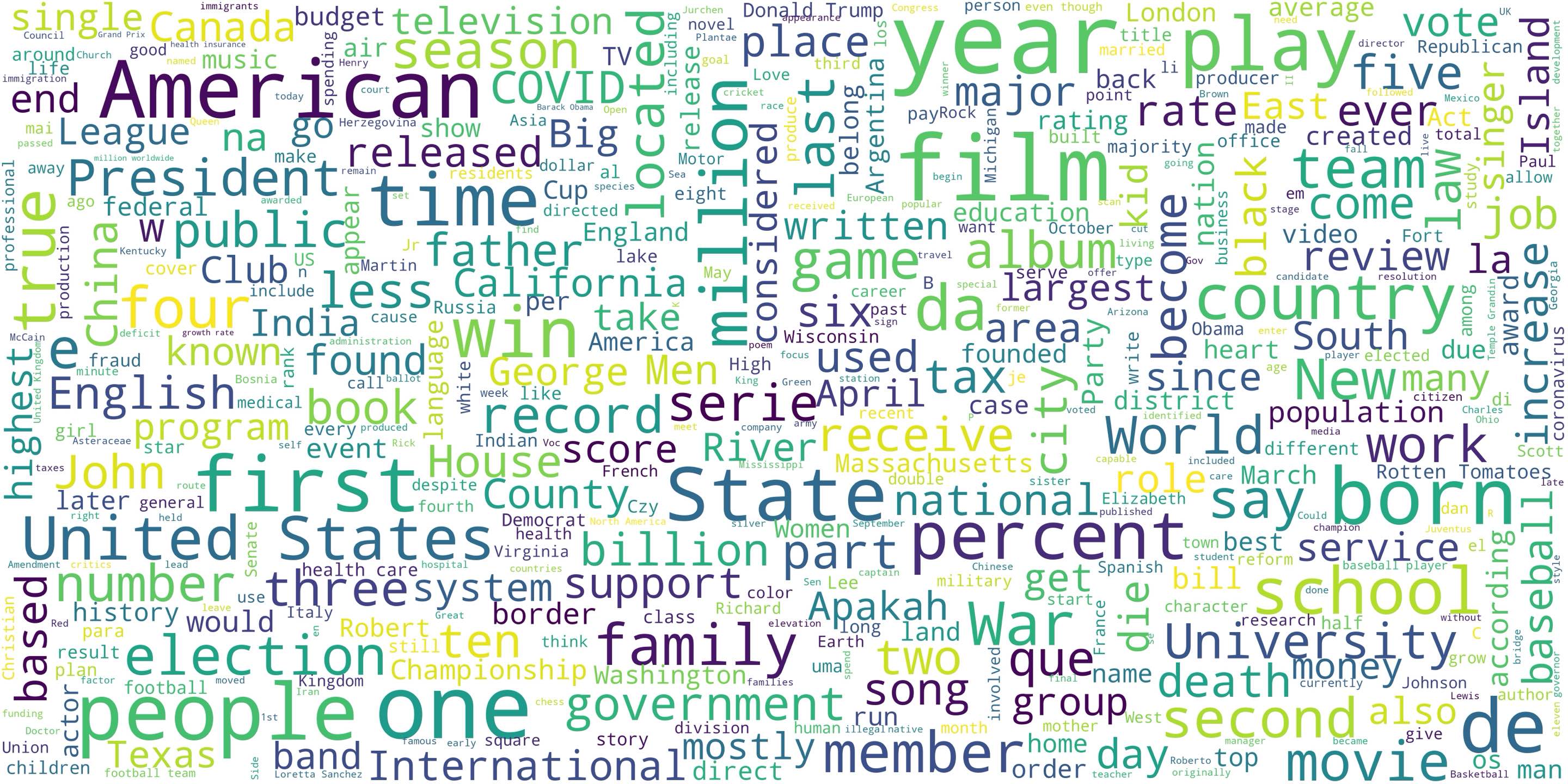}}
\label{fig:motivation_a}
\end{minipage}%
}%
\subfigure[\textsf{LLMFake}]{
\begin{minipage}[t]{0.33\linewidth}
\centering
\scalebox{0.85}{\includegraphics[width=6cm]{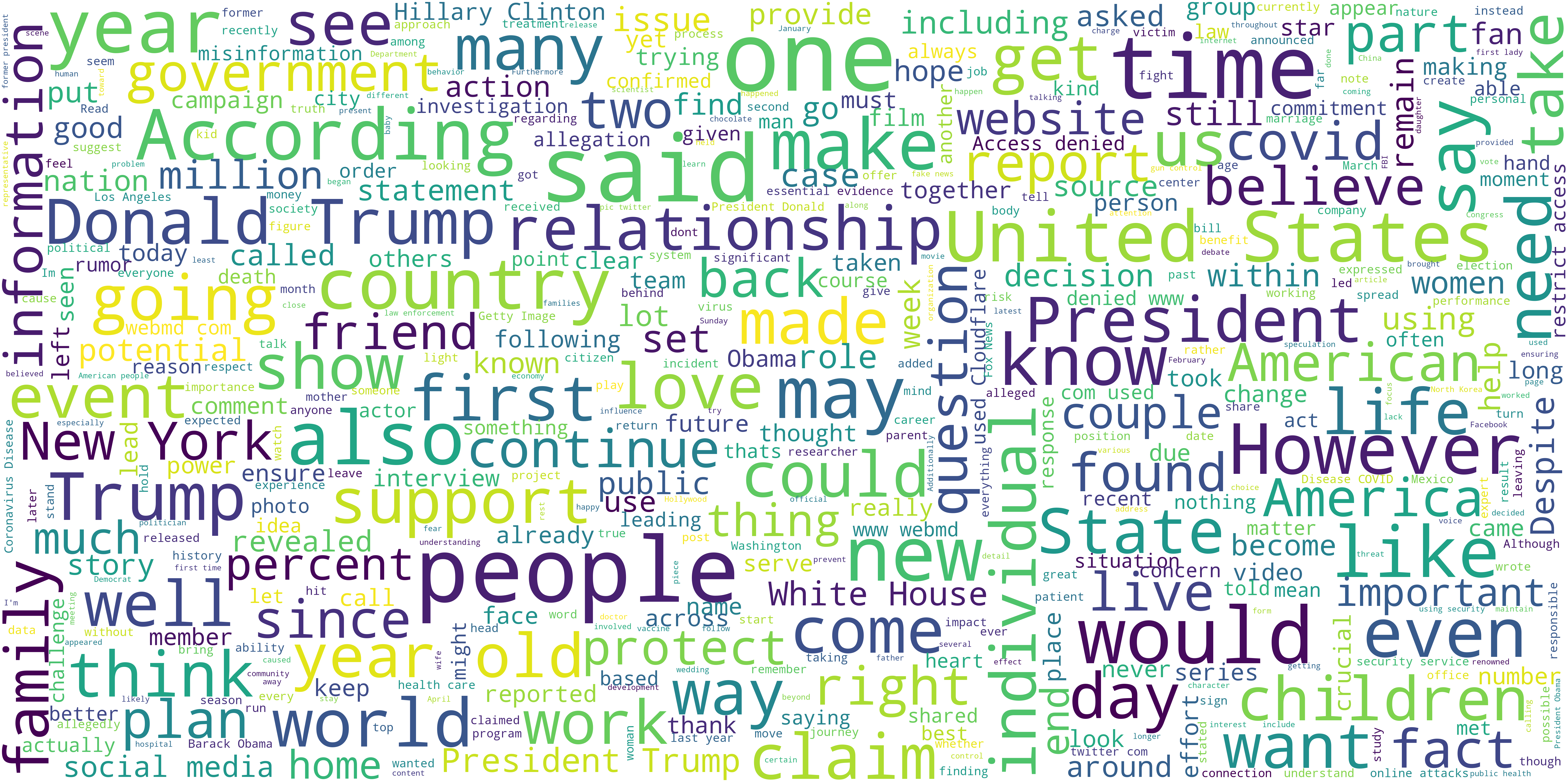}}
\label{fig:motivation_b}
\end{minipage}%
}%
\subfigure[\textbf{\textsc{Fact-Audit}}]{
\begin{minipage}[t]{0.33\linewidth}
\centering
\scalebox{0.85}{\includegraphics[width=6cm]{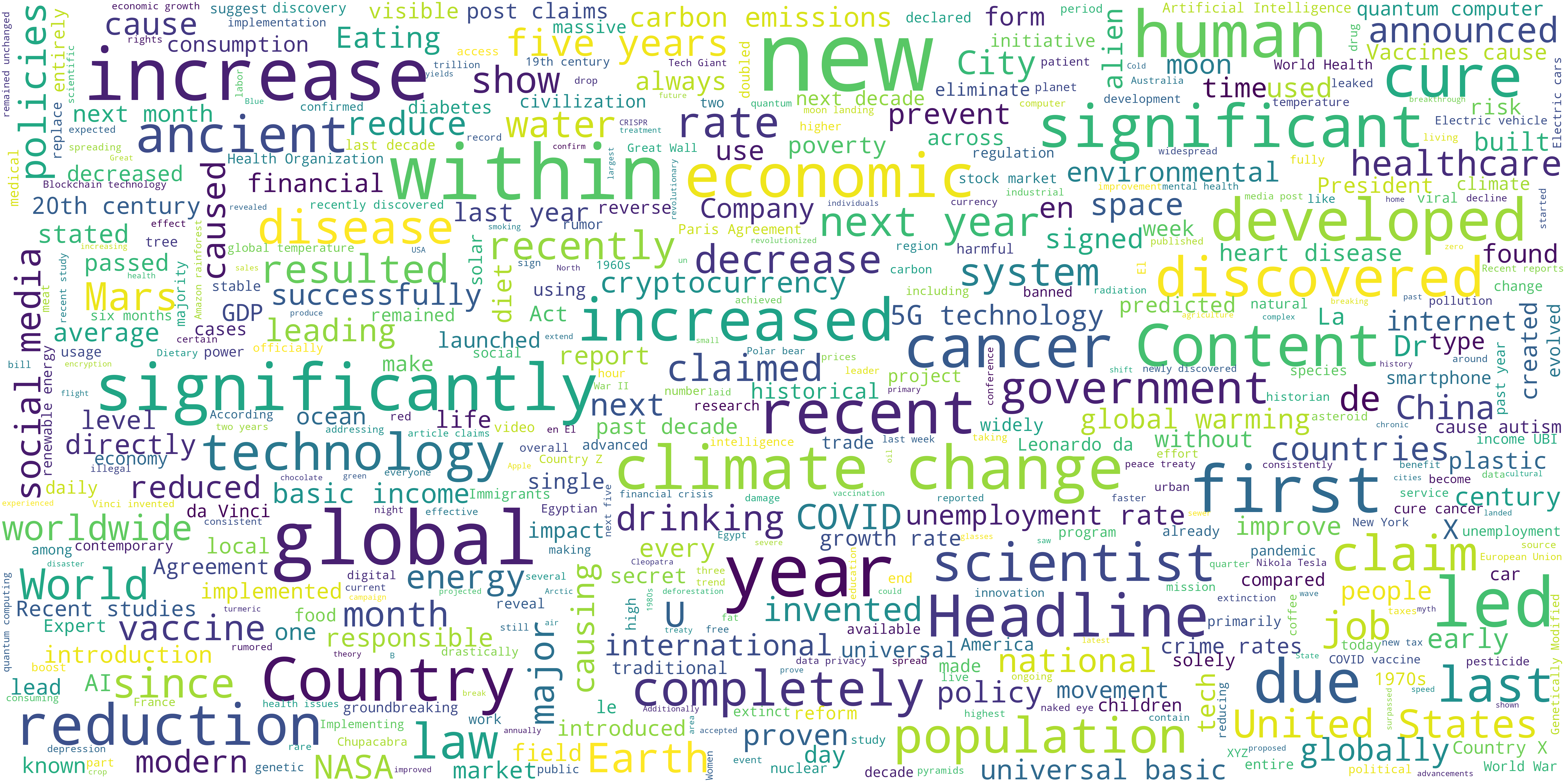}}
\label{fig:motivation_c}
\end{minipage}%
}
\centering
\vspace{-0.4cm}
\caption{Word clouds of the three benchmarks.}
\vspace{-0.2cm}
\label{fig:word_cloud}
\end{figure*}

The scores of human evaluation are shown in Table~\ref{tab:benchmark_comparison}. Note that the intra-class agreement score is 0.619 and the average Cohen's Kappa between any two expert annotators is 0.681. We can find that: 1) \textsc{Fact-Audit} has the lowest redundancy, indicating there is minimal repetition or unnecessary data due to the iterative probing. 2) \textsc{Fact-Audit} achieves the highest score in diversity benefitting from the adaptive updating, suggesting it includes a wider variety of test senarios. \textsf{LLMFake} consisting of LLM-generated misinformation also shows good diversity with a score of 2.17, while Pinocchio has the lowest diversity score of 1.94 since it is curated by human beings. We further provide the word clouds of the three benchmarks as shown in Figure~\ref{fig:word_cloud}. 3) \textsc{Fact-Audit} also tops in readability, indicating that its content is the easiest to understand. Pinocchio follows closely with 2.86, and \textsf{LLMFake} trails with a readability score of 2.43. 4) \textsc{Fact-Audit} leads in coverage as well, suggesting it comprehensively addresses the relevant fact-checking subjects or topics. Pinocchio that focuses on complex claims scores 2.14, and \textsf{LLMFake} with only fake news has the lowest coverage score of 1.65, indicating a narrower scope in addressing the intended fact-checking subjects. 5) The three benchmarks all perform reasonably well in fairness. 6) The scores for suitability are close across the datasets, but \textsc{Fact-Audit} slightly leads with 2.81, indicating its data is the most appropriate for automatic fact-checking evaluation of LLM auditing. 

\section{Detailed Performance by Test Modes} \label{DPTM}
Although the framework allows agents to autonomously determine the test modes for fact-checking data,
we provide the detailed results of the performance under different test modes as shown in Table~\ref{tab:test_mode_full}, to study the effect of different test modes.

\begin{table*}[t] \large
\centering
\scalebox{0.69}{
\begin{tabular}{@{}l|c|ccccccccc|ccc@{}}
\toprule
\multirow{2}{*}{Target LLM}  & \multirow{2}{*}{Test Mode} & \multicolumn{3}{c}{Complex Claim} & \multicolumn{3}{c}{Fake News} & \multicolumn{3}{c|}{Social Rumor}           & \multicolumn{3}{c}{Overall} \\ \cmidrule(l){3-5} \cmidrule(l){6-8} \cmidrule(l){9-11} \cmidrule(l){12-14}
                             &                            & \textit{IMR}       & \textit{JFR}       & \textit{Grade}     & \textit{IMR}      & \textit{JFR}      & \textit{Grade}   & \textit{IMR}   & \textit{JFR}   & \multicolumn{1}{c|}{\textit{Grade}} & \textit{IMR}     & \textit{JFR}     & \textit{Grade}   \\ \midrule
\multirow{3}{*}{Llama3.1-8B} & {[}claim{]}                & 76.88     & 24.38     & 3.13      & 61.39    & 21.94    & 3.87    & 63.75 & 21.25 & \multicolumn{1}{c|}{3.95}  & 68.80   & 22.87   & 3.56    \\
                             & {[}evidence{]}             & 44.38          & 14.79          & 5.08          & 26.41         & 11.54         & 6.27        & 44.07      & 13.33      & \multicolumn{1}{c|}{5.14}      & 38.16        & 13.33        & 5.50        \\
                             & {[}wisdom of crowds{]}     & 51.04     & 17.92     & 4.63      & 34.44    & 14.44    & 5.63    & 45.93 & 14.44 & \multicolumn{1}{c|}{4.88}  & 45.29   & 16.08   & 4.96    \\ \cmidrule(l){1-2} 
\multirow{3}{*}{Qwen2.5-7B}  & {[}claim{]}                & 49.67     & 14.33     & 4.72      & 38.89    & 12.22    & 5.28    & 53.96 & 12.08 & \multicolumn{1}{c|}{4.45}  & 48.86   & 12.76   & 4.74    \\
                             & {[}evidence{]}             & 20.62     & 10.83     & 6.65      & 8.21    & 4.62    & 7.14    & 44.76 & 4.29 & \multicolumn{1}{c|}{5.21}  & 20.83   & 7.31   & 6.45    \\
                             & {[}wisdom of crowds{]}     & 55.62     & 7.92     & 4.42      & 25.56    & 7.04    & 6.19    & 20.95 & 6.67 & \multicolumn{1}{c|}{6.78}  & 39.58   & 7.40   & 5.43    \\\cmidrule(l){1-2}
\multirow{3}{*}{GPT-4o}      & {[}claim{]}                & 25.83     & 20.00     & 6.02      & 23.64    & 15.45    & 6.02    & 16.67 & 11.67 & \multicolumn{1}{c|}{6.44}  & 23.05   & 16.67   & 6.11    \\
                             & {[}evidence{]}             & 16.33     & 12.67     & 6.69      & 6.46    & 5.00    & 7.24    & 11.39 & 10.56 & \multicolumn{1}{c|}{6.93}  & 10.61   & 8.77   & 7.00    \\
                             & {[}wisdom of crowds{]}     & 19.39     & 10.91     & 6.43      & 12.22    & 6.29    & 6.90    & 13.70 & 7.78 & \multicolumn{1}{c|}{6.73}  & 15.40   & 8.51   & 6.67    \\ \bottomrule
\end{tabular}}
\vspace{-0.2cm}
\caption{The fact-checking performance of three representative LLMs under three fixed test modes, respectively.}
\label{tab:test_mode_full}
\vspace{-0.4cm}
\end{table*}

\section{Diversity of Fact-checking Topics} \label{DFT}

\begin{figure}[]
  \includegraphics[width=\columnwidth]{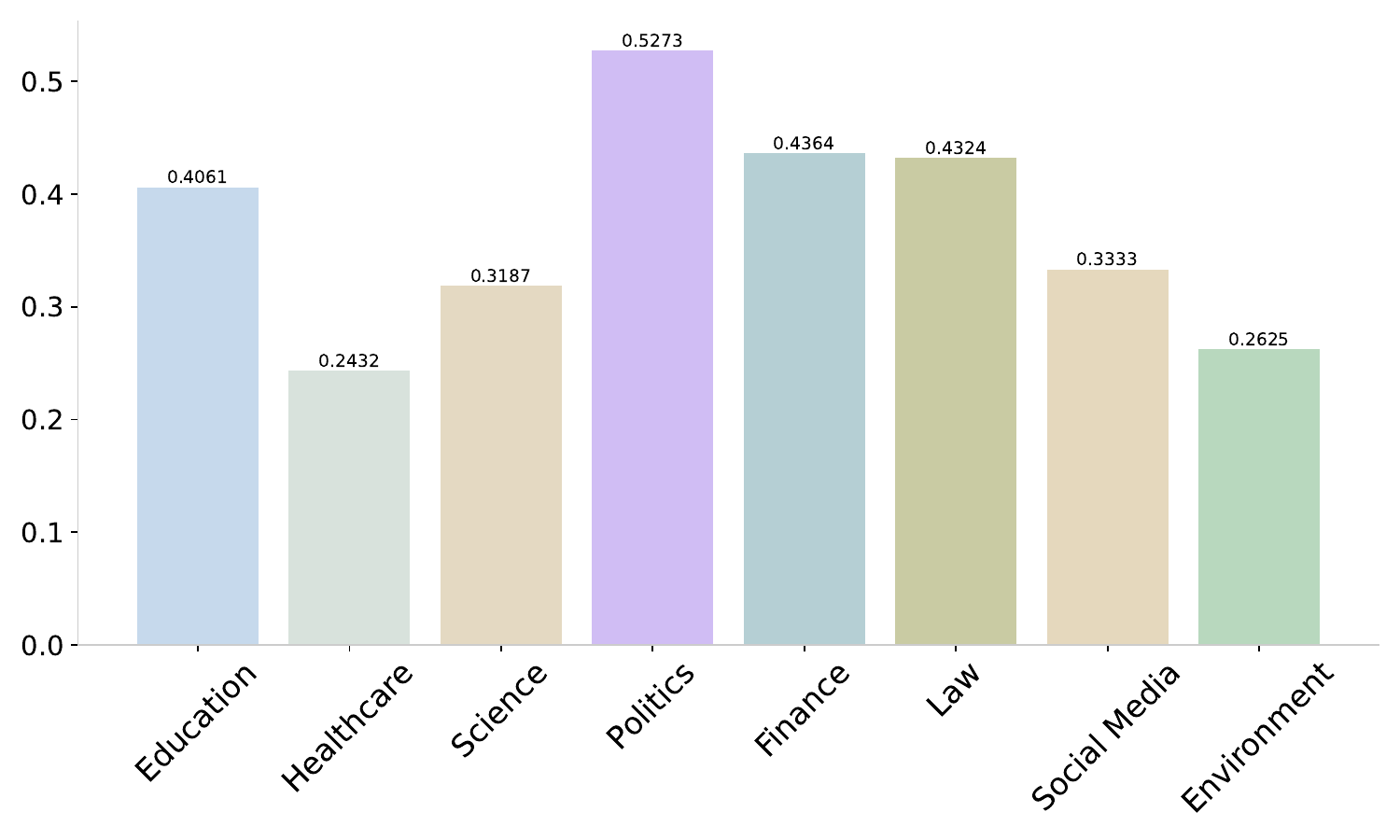}
  \vspace{-0.7cm}
  \caption{The \textit{IMR} performance on diverse fact-checking topics.}
  \label{fig:topics}
  \vspace{-0.3cm}
\end{figure}

Figure~\ref{fig:topics} allows us to analyze the performance of the target LLM across diverse topics using the Insight Mastery Rate (\textit{IMR}), by taking Qwen2.5-7B as an example. Here's a breakdown of the model's performance by topic:
1) Politics: With the highest \textit{IMR} of 52.73\%, the model struggles most in the political domain, likely due to the complexity and variability inherent in political content. 2) Finance and Law: Both areas show relatively high \textit{IMR}s of 43.64\% and 43.24\%, respectively, suggesting challenges in handling the intricate details and regulations prevalent in these fields.
3) Education: Here, the model achieves an \textit{IMR} of 40.61\%, indicating moderate performance that could benefit from further improvements, possibly due to the broad range of knowledge required in educational topics.
4) Science, Social Media, and Environment: These areas record better performance with \textit{IMR}s of 31.87\%, 33.33\%, and 26.25\%, respectively, with the model performing best in environmental topics. This suggests a stronger grasp in handling fact verification in these less politically or economically charged domains.
5) Healthcare: Excelling with the lowest \textit{IMR} of 24.32\%, this indicates that Qwen2.5-7B is particularly adept at processing and verifying facts in the healthcare sector, likely due to effective training or inherent capabilities in understanding medical contexts.

In summary, Qwen2.5-7B exhibits varied performance across different topics, facing the most significant challenges in politics, finance, and law, while showing strengths in healthcare, science, and environmental areas. This variability may point to differences in the volume and quality of training data or the model’s ability to process specific types of information within these domains. Areas of underperformance may benefit from targeted improvements in model training or enriched data inputs to enhance fact-checking in diverse topics. This also highlights the advantage of our framework that could investigate the target LLM dynamically in diverse aspects without the constraints of static data.

\section{Discussion of Adaptive Updating} \label{DAU}
As shown in Figure~\ref{fig:poor_scenario}, the most challenging test scenario ``Aggregated Statistical Reasoning'' was additionally excavated by \textsc{Fact-Audit} during the Adaptive Updating stage. This reaffirms that the Adaptive Updating stage could explore more diverse and challenging test data as the research statement claimed in \S\ref{method}. Besides, the Adaptive Updating stage totally designed 4, 3, 1 new challenging test scenarios where the \textit{IMR} performance of target LLM is worse than the averaged performance, for the target LLM Qwen2.5-72B in Complex Claim, Fake News, and Social Rumor, respectively.


\section{More Example Illustrations}
\label{more_case}

\begin{figure}[t]
    \centering
    \includegraphics[width=1\linewidth]{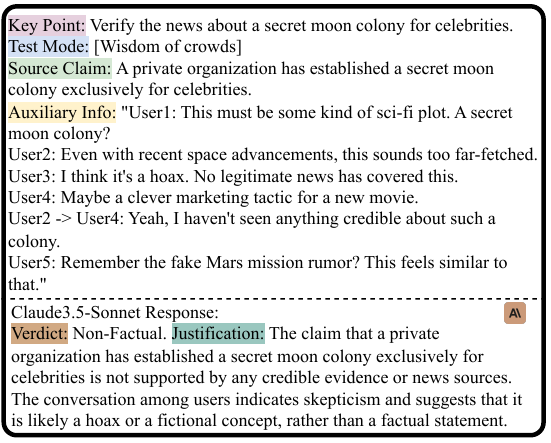}
    \vspace{-0.7cm}
    \caption{Example of cases with correct verdict prediction yet poor justification production of Claude3.5-Sonnet for fact-checking the fake news.}
    \label{fig:case3}
    \vspace{-0.3cm}
\end{figure}

\begin{figure}[t]
    \centering
    \includegraphics[width=1\linewidth]{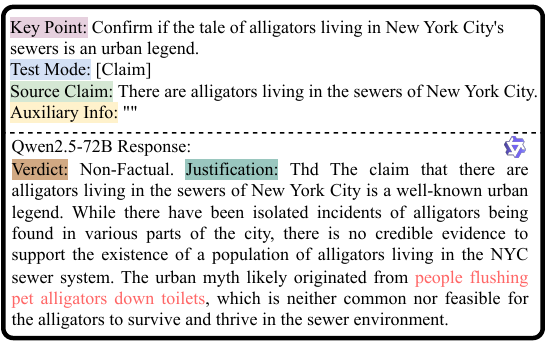}
    \vspace{-0.7cm}
    \caption{Example of cases with correct verdict prediction yet poor justification production of Qwen2.5-72B for fact-checking the social rumor.}
    \label{fig:case1}
    \vspace{-0.3cm}
\end{figure}

Since we have provided a case about the GPT-4o response under the [evidence] mode on the complex claim in \S\ref{case_study}, here we further provide two cases: 1) one about the Claude3.5-Sonnet response under the [wisdom of crowds] mode on the fake news, as shown in Figure~\ref{fig:case3}; 2) one about the Qwen2.5-72B response under the [claim] mode on the social rumor, as shown in Figure~\ref{fig:case1}.

For the case in Figure~\ref{fig:case}, we provide the output evaluation given by the Evaluator agent for better explainability and reliability as follows:

``\textit{The assistant's response begins by declaring the claim about bamboo's growth rate as "Factual," which is accurate given the context of the question. It references the same credible sources as the reference response — Wikipedia and Guinness World Records — which state that some bamboo species can indeed grow up to 36 inches (91 cm) in a day under ideal conditions. This aligns with the claim that bamboo can grow up to 35 inches, as 35 inches is slightly less than 36 inches and falls near the same growth range.}

\textit{However, the assistant makes a slight error in its justification. It states that bamboo can grow "up to 35 inches (91 cm)," but 35 inches is not equivalent to 91 cm; 91 cm is equivalent to 36 inches. This inconsistency introduces confusion. Furthermore, the assistant does not adequately address the importance of the specific conditions needed for such growth, which the reference response highlights. These are critical points that temper the general claim about bamboo's growth rates, suggesting that the claim misses its completeness.}

\textit{In terms of soundness and relevance, the assistant's justification is on topic as it discusses both the claim's veracity and the conditions affecting growth. However, due to the critical inaccuracies and lack of nuance regarding the context from the reference response, the assistant's quality suffers.}

\textit{Overall, while the assistant's conclusion is correct, the justification is flawed and lacks the necessary depth to provide a thorough understanding of the climatic conditions that affect bamboo growth.}

\textit{Rating: [[2]]}''.

For the case in Figure~\ref{fig:case3}, we provide the output evaluation given by the Evaluator agent for better explainability and reliability as follows:

``
\textit{The assistant's response correctly identifies the claim as non-factual and rightly states that there is no credible evidence to support the establishment of a secret moon colony for celebrities. This aligns well with the reference response, which also concludes that the claim lacks verification and is considered improbable.}

\textit{However, there are a few areas where the assistant's justification could be improved:}

\textit{1. **Completeness**: While the assistant mentions skepticism among users, it doesn’t elaborate on the specifics of that skepticism or relate it to the broader context of space travel challenges. The reference response includes a detailed analysis of the skepticism expressed by multiple users and draws a parallel to debunked rumors, providing a richer context for understanding why the claim is implausible.}

\textit{2. **Relevance**: The assistant's response is relevant but lacks depth in discussing why the claim is likely a hoax or a plot from a sci-fi narrative, as mentioned in the reference response. It also misses the aspect of comparing this claim to the fake Mars mission rumor, which adds an additional layer of understanding.}

\textit{Overall, while the assistant's answer is correct, the justification provided is not as comprehensive or informative as the reference response, which presents a more robust analysis of the situation.}

\textit{Rating: [[3]]}
''.

For the case in Figure~\ref{fig:case1}, we provide the output evaluation given by the Evaluator agent for better explainability and reliability as follows:

``
\textit{The assistant’s response labels the claim as "Non-Factual," which aligns with the classification made in the reference response. However, when evaluating the correctness and completeness of the justification, several points must be addressed:}

\textit{1. **Correctness**: The assistant correctly identifies the alligator tale as an urban legend. However, the statement about isolated incidents is somewhat misleading. While there have been occasional reports of alligators being found in NYC, these do not substantiate the claim of a thriving population in the sewers.}

\textit{2. **Soundness**: The assistant mentions the origin of the myth regarding flushing pet alligators. This point provides context but lacks clarity on how this contributed to the legend. The explanation could be refined to better emphasize that these incidents do not lead to a sustainable population in the sewer system.}

\textit{3. **Relevance**: The assistant's response is relevant to the question. However, the mention of isolated incidents, without a clear explanation that these do not contribute to a rabid alligator population, might distract from the core message.}

\textit{4. **Completeness**: The response provides most of the necessary information but lacks a direct reference to expert consensus or evidence debunking the myth, which is an important aspect of the justification.}

\textit{In summary, while the assistant's response is mostly accurate, it contains minor inaccuracies regarding the significance of isolated incidents and lacks stronger supporting evidence against the myth, similar to the depth present in the reference response.}

\textit{Rating: [[3]]}
''.


\section{Quick Proof of Convergence}
The convergence of our framework is ensured with $\pi(\Theta_{i+1}|\Theta_i,\mathcal{M})$ that $\Theta_{i+1}$ is more likely to contain the fact-checking limitations of the target LLM, which can be formulated as:
{
\setlength{\abovedisplayskip}{0.1cm}
\setlength{\belowdisplayskip}{0.1cm}
\begin{equation}\small
    \mathbb{E}_{q_{i+1}(x)}\left[\mathcal{F}_\alpha(x)p(x)\right]\geq \mathbb{E}_{q_{i}(x)}\left[\mathcal{F}_\alpha(x)p(x)\right].
\end{equation}
It further derives that the variance of \Cref{eq:obj} keeps decreasing during iterations:
{
\setlength{\abovedisplayskip}{0.1cm}
\setlength{\belowdisplayskip}{0.1cm}
\begin{equation}\small
    \mathop{Var}_{\quad q_{i+1}}\left[\mathcal{F}_\alpha(x)\frac{p(x)}{q_{i+1}(x)}\right]\leq \mathop{Var}_{\quad q_{i}}\left[\mathcal{F}_\alpha(x)\frac{p(x)}{q_{i}(x)}\right].
\end{equation}}
In addition, since we start from $q_0(x) = p(x)$, there is $\mathop{Var}\nolimits_{q_{i+1}}\leq\mathop{Var}\nolimits_{q_i}\leq \cdots\leq \mathop{Var}\nolimits_{p}$, which means that our method converges faster than direct sampling from $p(x)$, with the convergence speed increasing in each iteration. This further validates the reliability of our proposed framework.

\section{Ratio of Poor Justification in Bad Cases}

\begin{table}[t] \large
\centering
\scalebox{0.65}{
\begin{tabular}{@{}l|ccc|c@{}}
\toprule
\normalsize{Model}             & \normalsize{Complex Claim} & \normalsize{Fake News} & \normalsize{Social Rumor} & \normalsize{Overall} \\ \midrule
\normalsize{Mistral-7B}         & 42.55         & 41.22     & 66.93        & 42.60   \\
\normalsize{Llama2-7B}         & 42.41         & 55.56     & 42.42        & 45.47   \\
\normalsize{Llama2-13B}        & 32.99         & 30.12     & 42.57        & 34.05   \\
\normalsize{Llama3-8B}         & 30.37         & 51.23     & 41.44        & 40.30   \\
\normalsize{Llama3.1-8B}       & 38.43         & 35.11     & 27.00        & 35.27   \\
\normalsize{Llama3.1-70B}      & 34.22         & 47.50     & 26.09        & 36.34   \\
\normalsize{Qwen2.5-7B}        & 25.00         & 38.10     & 14.77        & 25.62   \\
\normalsize{Qwen2.5-72B}       & 24.53         & 14.00     & 25.00        & 21.88   \\
\normalsize{GLM4-9B}          & 31.03         & 27.10     & 30.48        & 29.49   \\
\normalsize{Gemma2-9B}     & 67.20         & 79.19     & 52.94        & 67.43   \\
\normalsize{Gemini-Pro}       & 39.31         & 29.69     & 17.39        & 31.65   \\
\normalsize{Claude3.5-Sonnet} & 28.66         & 15.56     & 17.95        & 24.48   \\
\normalsize{GPT-4o} & 30.89         & 27.75     & 13.45        & 72.30   \\\bottomrule
\end{tabular}}
\caption{The JFR/IMR ratio of poor justification for correct verdict prediction in bad cases.}
\label{jfr/imr}
\end{table}

Table~\ref{jfr/imr} demonstrates the ratio of cases with poor justifications yet correct verdict predictions and the total bad cases with rating grades below 4.0.

\section{Discussion of Potential Bias}
Previous literature~\cite{petroni2019language, jiang2020can} has shown that LLMs store factual knowledge and function as knowledge bases, which aids knowledge-intensive tasks like truthful question answering and fact-checking~\cite{roberts2020much, lin2022amif, lin2022truthfulqa, pan2023fact}.
In this work, we focus on GPT-4o as the agent controller, as it is widely regarded as one of the most capable LLMs currently available. While LLM-as-a-Judge evaluation introduces potential bias in most generation tasks, particularly in quality evaluation, this bias is analogous to the inherent cognitive bias observed in human judges. Similarly, LLM judges, like GPT-4o, may exhibit biases due to their limited stored knowledge. Therefore, the generated fact-checking data may not cover all the real-world scenarios. That's also why we propose such an adaptive multi-agent framework for dynamic fact-checking evaluation. However, this does not hinder our ability to address the research questions posed in this paper, as we have implemented a series of measures to mitigate these biases. Nevertheless, developing more reliable LLM judges remains a key challenge for future research. In complex scenarios and cross-domain fact-checking applications, LLM judges hold significant potential for advancing this field. In future research, we aim to integrate more advanced agent configurations as LLMs continue to evolve, replacing the current dominant GPT-4o. Additionally, we plan to incorporate human-in-the-loop procedures to enhance the reliability and robustness of our evaluation framework. This will serve as a crucial direction for further exploration.





\end{document}